\documentclass{article}

\PassOptionsToPackage{numbers, compress}{natbib}

\usepackage[final]{neurips_2022}

\usepackage[utf8]{inputenc} 
\usepackage[T1]{fontenc}    
\usepackage{hyperref}       
\usepackage{url}            
\usepackage{booktabs}       
\usepackage{amsfonts}       
\usepackage{nicefrac}       
\usepackage{microtype}      
\usepackage{xcolor}         

\definecolor{darkgreen}{rgb}{0,0.5,0}
\hypersetup{colorlinks=true, citecolor=darkgreen}
\usepackage{enumitem} 
\usepackage{subcaption} 
\usepackage{adjustbox} 
\usepackage{multirow} 
\usepackage{amsmath} 
\usepackage{bbm} 
\usepackage{mathtools} 
\usepackage{wrapfig} 

\usepackage{colortbl} 
\usepackage{arydshln} 

\usepackage{pifont} 

\usepackage{tabularx}
\newcolumntype{L}[1]{>{\raggedright\let\newline\\\arraybackslash\hspace{0pt}}m{#1}}
\newcolumntype{C}[1]{>{\centering\let\newline\\\arraybackslash\hspace{0pt}}m{#1}}

\usepackage{pdfpages}

\title{
ALIFE: Adaptive Logit Regularizer and Feature Replay for Incremental Semantic Segmentation
}

\author{
Youngmin Oh
\qquad
Donghyeon Baek
\qquad
Bumsub Ham\thanks{Corresponding author.}
\\
School of Electrical and Electronic Engineering, Yonsei University
\\
\url{https://cvlab.yonsei.ac.kr/projects/ALIFE}
}

\begin{document}

\maketitle

\begin{abstract}
We address the problem of incremental semantic segmentation~(ISS) recognizing novel object/stuff categories continually without forgetting previous ones that have been learned. The catastrophic forgetting problem is particularly severe in ISS, since pixel-level ground-truth labels are available only for the novel categories at training time. To address the problem, regularization-based methods exploit probability calibration techniques to learn semantic information from unlabeled pixels. While such techniques are effective, there is still a lack of theoretical understanding of them. Replay-based methods propose to memorize a small set of images for previous categories. They achieve state-of-the-art performance at the cost of large memory footprint. We propose in this paper a novel ISS method, dubbed ALIFE, that provides a better compromise between accuracy and efficiency. To this end, we first show an in-depth analysis on the calibration techniques to better understand the effects on ISS. Based on this, we then introduce an adaptive logit regularizer~(ALI) that enables our model to better learn new categories, while retaining knowledge for previous ones. We also present a feature replay scheme that memorizes features, instead of images directly, in order to reduce memory requirements significantly. Since a feature extractor is changed continually, memorized features should also be updated at every incremental stage. To handle this, we introduce category-specific rotation matrices updating the features for each category separately. We demonstrate the effectiveness of our approach with extensive experiments on standard ISS benchmarks, and show that our method achieves a better trade-off in terms of accuracy and efficiency.
\end{abstract}

\vspace{-.4cm}
\section{Introduction}
Humans are capable of learning new concepts continually, while preserving or even improving previously acquired knowledge. Artificial neural networks are, however, prone to forget the knowledge they have learned if being trained with samples for new object/scene categories alone. The reason for this problem, so-called catastrophic forgetting~\cite{french1999catastrophic,mccloskey1989catastrophic}, is that parameters of neural networks change abruptly to handle new categories without accessing training samples for previous categories. A straightforward way to alleviate the problem is to re-train a model with training examples for entire categories observed so far, which is however computationally demanding. 

Incremental learning is an alternative approach to learning new categories continuously without re-training on the entire dataset. While many methods have been proposed for incremental classification~\cite{chaudhry2018riemannian,kirkpatrick2017overcoming,li2017learning,lopez2017gradient,rebuffi2017icarl,wu2019large}, a few attempts explore incremental semantic segmentation~(ISS), where training images for new categories are partially labeled to reduce the cost for manual annotation. That is, pixels for new categories are labeled only, while remaining ones are marked as unknown. The unknown regions should be considered separately, since they could contain previous categories along with ones that would be seen in the future.

Current ISS methods can be categorized into regularization-based~\cite{MIB20cer,PLOP21dou,SDR21mic} and replay-based approaches~\cite{SSUL21cha,RECALL21mar}. Regularization-based methods typically exploit knowledge distillation~(KD)~\cite{hinton2014distilling} to preserve the discriminative ability of ISS models for previous categories. In particular, the seminal work of~\cite{MIB20cer} alleviates the semantic shift of unknown regions. Specifically, it introduces calibrated cross-entropy~(CCE) and calibrated KD~(CKD) terms to learn semantic information from unknown regions. Although CCE and CKD terms have shown the effectiveness on ISS methods~\cite{MIB20cer,PLOP21dou,SDR21mic}, there is still a lack of theoretical understanding. Replay-based methods propose to use a replay buffer consisting of web-crawled or previously seen images. The replay buffer provides rich information storing the knowledge for previous categories, but it incurs large memory footprint, particularly for the task of semantic segmentation that typically adopts high-resolution images.

\begin{figure*}[t]
	\centering
	\begin{subfigure}{0.335\textwidth}
 		\vspace{-.25cm}
		\includegraphics[width=\linewidth]{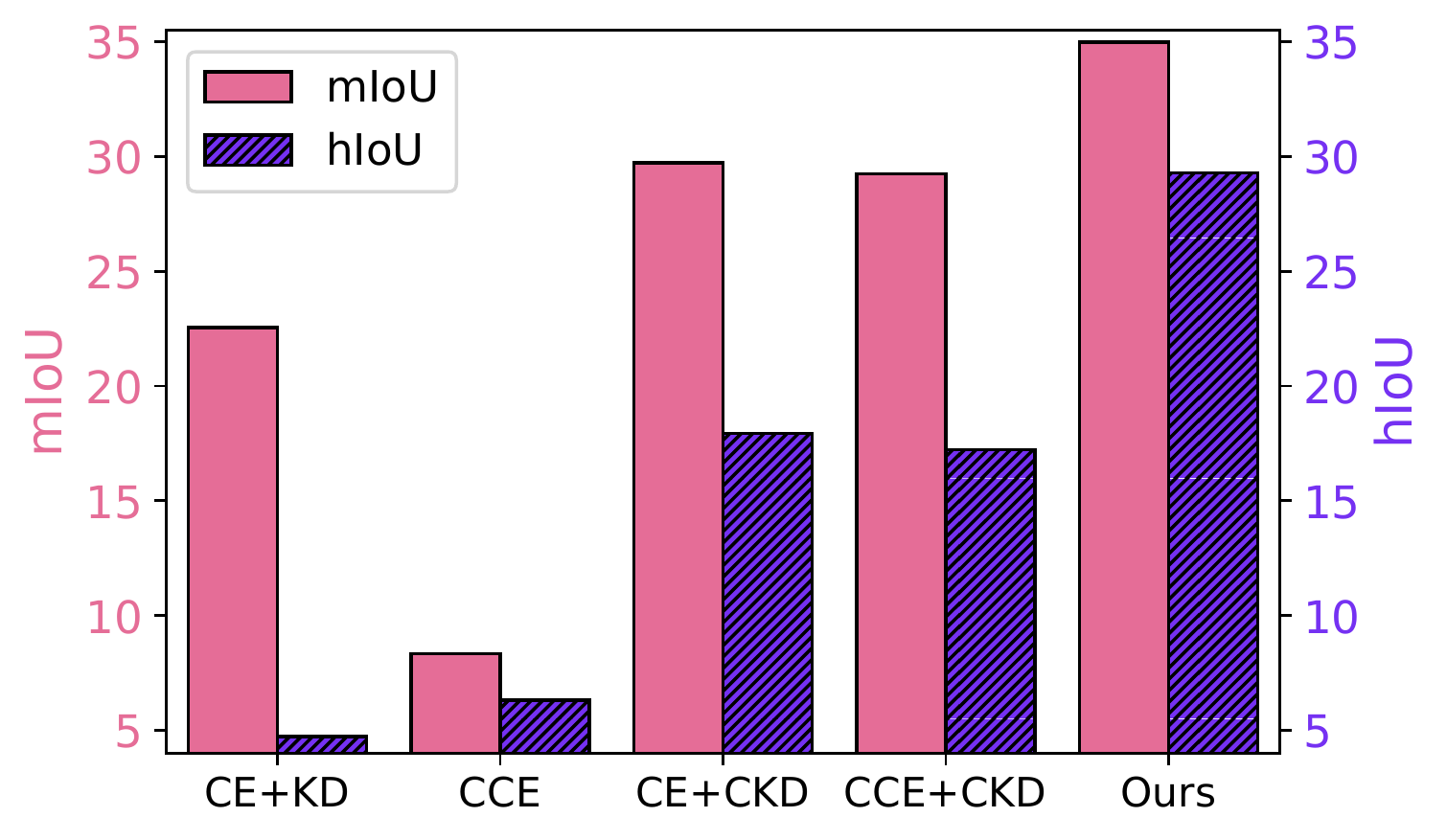}
	\end{subfigure}
	\begin{subfigure}{0.31\textwidth}
		\includegraphics[width=\linewidth]{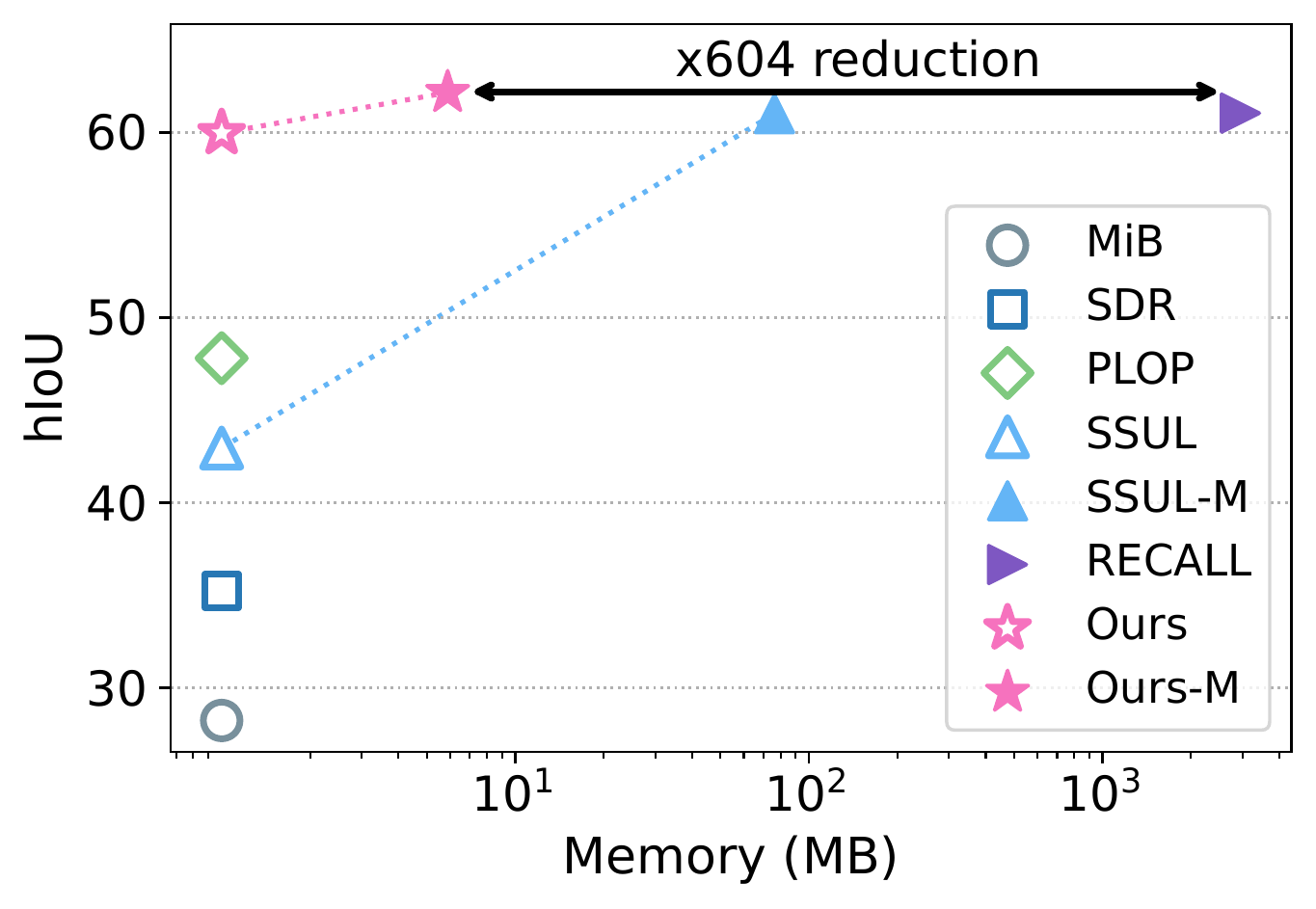}
	\end{subfigure}
	\begin{subfigure}{0.31\textwidth}
		\includegraphics[width=\linewidth]{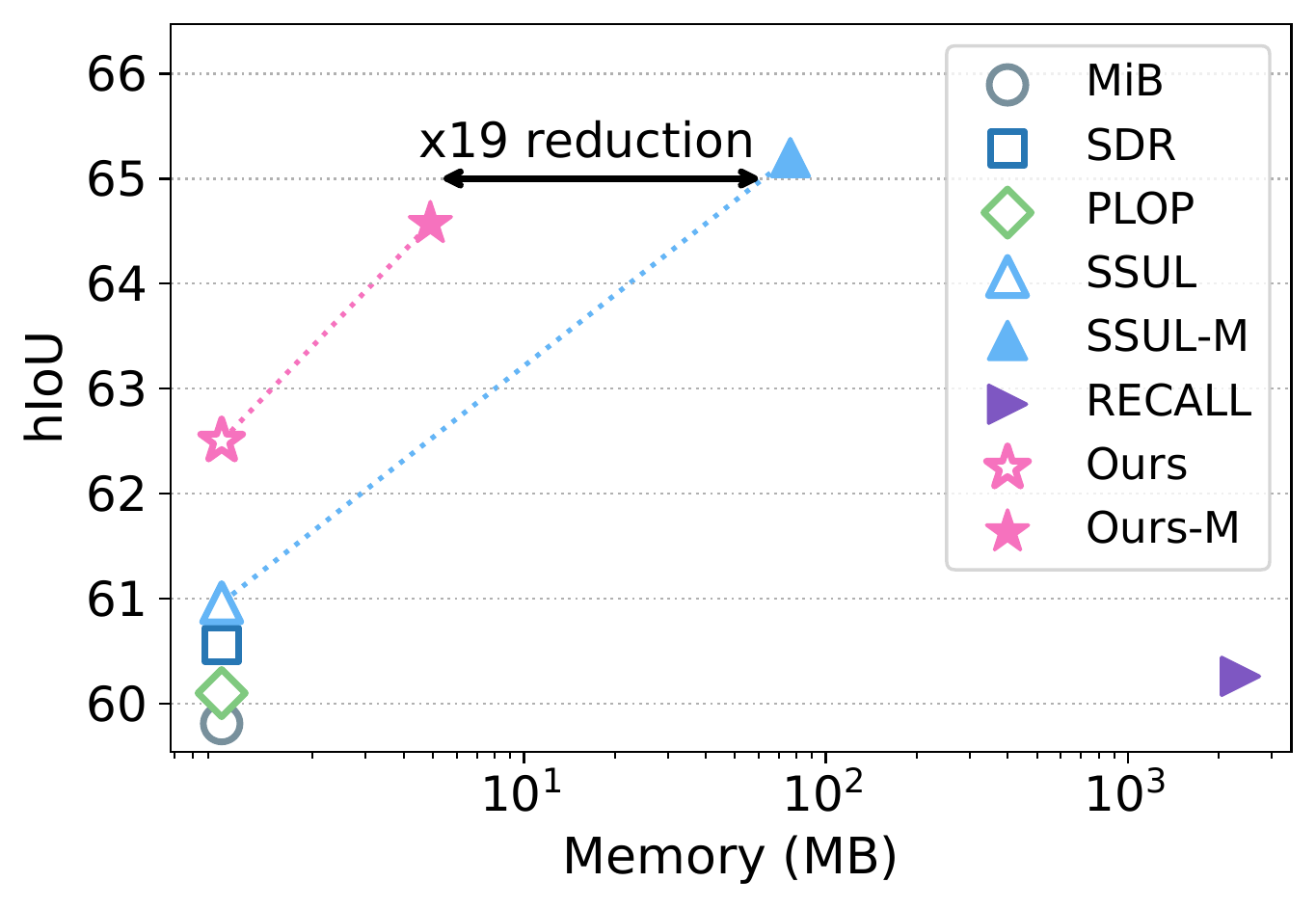}
	\end{subfigure}
	
	\begin{minipage}{0.34\textwidth}
		\centering
		\small{(a) IoU scores on~ADE20K~\cite{ADE17zhou}.}
	\end{minipage}
	\begin{minipage}{0.62\textwidth}
		\centering
		\small{(b) hIoU-memory comparison on PASCAL VOC~\cite{VOC10ever}.}		
	\end{minipage}
	\caption{(a)~Quantitative comparison of intersection-of-union~(IoU) scores on ADE20K. Each model learns 50 novel categories after learning 100 categories. To be specific, there is a total of 5 incremental stages, and each model learns 10 new categories at every incremental stage. mIoU: an average IoU score for 150 categories. hIoU: a harmonic mean between two average IoU scores for previous~(100) and novel~(50) categories.~(b)~Quantitative comparison of hIoU and memory requirement on PASCAL VOC. (From left to right) Each method learns 1 and 5 novel categories after learning 20 and 16 categories, respectively. Ours-M memorizes 1K features for each previous category. For RECALL~\cite{RECALL21mar} and SSUL-M~\cite{SSUL21cha}, we consider the size of memory required to store images only, and discard that for corresponding labels or saliency maps. Best viewed in color.}
	\label{fig:teaser}
\end{figure*}

In this paper, we present a novel ISS method, dubbed ALIFE, that 1)~alleviates catastrophic forgetting and 2)~reduces memory requirements. For the first aspect, we analyze gradients of CCE and CKD terms for better understanding the effectiveness on catastrophic forgetting, and make the following observations:~(1)~For unknown regions, CCE reduces logit values of new categories, which is crucial for preventing overfitting to the new categories. However, it always raises logit values of all previous categories, without considering whether predictions for those regions are correct or not, which lessens the discriminative power for previous categories. (2)~CKD makes it difficult to distinguish new categories from a background one. Motivated by these observations, we introduce an adaptive logit regularizer~(ALI) that enables better learning new categories and alleviating catastrophic forgetting for previous ones~(Fig.~\ref{fig:teaser}(a)). For the second aspect, we propose to exploit latent features for replaying, reducing memory footprint and avoiding data privacy issues~(\emph{e.g.}, in medical imaging~\cite{shokri2015privacy}). In contrast to the replay buffer storing images directly, our approach to using the features updates them at every incremental stage, as a feature extractor is also updated continually at training time. Specifically, we exploit category-specific rotation matrices using the Cayley transform. Rotating latent features is computationally efficient, while maintaining the relations between the features. We demonstrate that exploiting ALI with memorizing features achieves a better trade-off in terms of accuracy and efficiency~(Fig.~\ref{fig:teaser}(b)). Our main contributions are summarized as follows:
\begin{itemize}[leftmargin=0.2in]
	\item We show an in-depth analysis of probability calibration methods widely used for ISS~\cite{MIB20cer,PLOP21dou,SDR21mic}, and introduce ALI that enables our model to better learn new categories, while maintaining the knowledge for previous categories.
	\item We present a novel replay strategy using category-specific rotation matrices, which helps to alleviate catastrophic forgetting for previous categories with much less memory requirements than replaying raw images.
	\item Extensive experiments demonstrate the effectiveness of our approach to using ALI and replaying features with rotation matrices. We set a new state of the art on standard ISS benchmarks~\cite{VOC10ever,ADE17zhou}.
\end{itemize}

\section{Related work}
\vspace{-.2cm}
Many incremental learning approaches have been introduced to preserve knowledge for previous categories, while recognizing new ones~\cite{mermillod2013stability}. They can be categorized into task-incremental and class-incremental methods. Task-incremental methods~\cite{kirkpatrick2017overcoming,li2017learning,lopez2017gradient} treat learning a new set of categories as a new task. With a perfect task identifier, they classify each task separately at test time. The perfect identifier is however often not available in practice~\cite{prabhu2020gdumb}. On the contrary, class-incremental methods~\cite{chaudhry2018riemannian,POD20dou,rebuffi2017icarl,wu2019large} attempt to recognize all categories observed so far at once without the task identifier, being more practical. In the following, we describe class-incremental methods pertinent to ours. 

\vspace{-.3cm}
\paragraph{Image classification.}
Incremental methods typically adopt a KD technique~\cite{hinton2014distilling} in order to encourage a current model to imitate softmax probabilities~\cite{castro2018end,rebuffi2017icarl} or intermediate representations~\cite{POD20dou,hou2019learning} obtained from a previous one, retaining the discriminability for previous categories. Another line of works focuses on preventing a classifier from overfitting to new categories, by correcting biased weights of a classifier with a post-processing~\cite{wu2019large,zhao2020maintaining} or normalizing a classifier~\cite{hou2019learning}. Similarly, our ALI is beneficial to alleviating the overfitting problem and maintaining the discriminative power for previous categories, but differs in that it adaptively regularizes logit values during training. In addition, we exploit latent features for replaying, while all the aforementioned methods rely on an image replay strategy that entails practical issues such as data privacy and large memory requirements. To avoid these problems, recent works employ a pseudo replay scheme that synthesizes images~\cite{kemker2017fearnet,ostapenko2019learning,shin2017continual} or latent features~\cite{xiang2019incremental} of previous categories. However, they often require a generator~\cite{goodfellow2014generative}, which should also be trained incrementally. Instead of using raw images or synthesizing features, the work of~\cite{FAN20isc} proposes to store latent features. It trains a feature adaptation network~(FAN) mapping the latent features into a new feature space. The FAN might be sub-optimal, since the same network applies to all features without differentiating categories. Adopting individual FANs for each category might address the problem, but this requires high computational cost and large memory footprint. Differently, we propose to train category-specific rotation matrices. This is more efficient and accurate, as a rotation matrix is light-weight and features of each category are updated separately. 

\vspace{-.3cm}
\paragraph{Semantic segmentation.}
A relatively few methods address the task of ISS, which can further be classified into regularization-based and replay-based approaches. The first approaches~\cite{MIB20cer,PLOP21dou,SDR21mic} focus on preserving knowledge for previous categories without using an experience replay. For example, MiB~\cite{MIB20cer} proposes to consider the semantic shift of a background category in ISS, and introduces CCE and CKD terms using probability calibration techniques. While the calibration techniques are intuitive and effective, there are still lack of theoretical explanations. We provide a gradient analysis on CCE and CKD, and point out that CCE and CKD could disturb preserving knowledge for previous categories and discriminating new categories from the background category, respectively. SDR~\cite{SDR21mic} and PLOP~\cite{PLOP21dou} propose to further regularize a current model in a latent feature space. Specifically, SDR minimizes distances between features of the same category~\cite{chopra2005learning}, while PLOP distills intermediate features using a multi-scale strip pooling technique~\cite{POD20dou,hou2020strip}. The second approaches~\cite{SSUL21cha,RECALL21mar} rely on an image replay strategy. RECALL~\cite{RECALL21mar} exploits web-crawled images with pseudo labels at each incremental stage, while SSUL~\cite{SSUL21cha} maintains a small set of images for previous categories, together with corresponding ground-truth masks. Both methods show state-of-the-art results at the cost of large memory footprint. Our approach differs in two aspects: (1) We exploit latent features for replaying, which reduces the size of required memory significantly; (2) Our method updates a feature extractor continually, and thus it is more flexible than RECALL and SSUL freezing the extractor.

\section{Method} \label{sec:3}
\vspace{-.2cm}
In this section, we first describe the ISS task briefly~(Sec.~\ref{sec:3.1}), and then present our approach consisting of three steps at every incremental stage. Specifically, we introduce ALI based on an in-depth analysis for CCE and CKD to train a ISS model~(Sec.~\ref{sec:3.2}) and present a feature replay scheme using category-specific rotation matrices~(Sec.\ref{sec:3.3}) to fine-tune a classifier using memorized features~(Sec.~\ref{sec:3.4}). Note that the last two steps are optional. Please refer to the supplementary material for detailed derivations, more analysis, and a pseudo code of our approach.

\subsection{Problem statement} \label{sec:3.1}
Following the common practice~\cite{MIB20cer,SSUL21cha,PLOP21dou,RECALL21mar,SDR21mic}, we consider a series of~$T$ learning stages. Each stage~$t$ has its own dataset,~$D^t=\{(x_i,y_i)\}^{n^{t}}_{i=1}$ of size~$n^{t}$, where~$x$ and~$y$ are an image and a corresponding ground-truth mask, respectively. For incremental stages~(\emph{i.e.},~$t>1$), we have two disjoint sets,~$C^{t}_\mathrm{prev}$ and~$C^{t}_\mathrm{new}$, for previously learned and novel categories, respectively. Note that ground-truth masks are labeled only for the categories of~$C^{t}_\mathrm{new}$. Formally, we denote by~$y(\mathbf{p})$ a ground-truth label at position~$\mathbf{p}$, and define labeled regions as follows:
\begin{equation}
	\mathcal{R}^{t}_\mathrm{new} = \bigcup_{c \in C^{t}_\mathrm{new}} \mathcal{R}_c,
\end{equation}
where~$\mathcal{R}_c$ is a set of locations labeled as a category~$c$, that is,~$\mathcal{R}_c = \{\mathbf{p} \mid y(\mathbf{p})=c\}$.

Our goal is to train a model that recognizes all categories observed so far,~$C^{t}_\mathrm{all}$, the union of the disjoint sets,~\emph{i.e.},~$C^{t}_\mathrm{all} = C^{t}_\mathrm{prev} \cup C^{t}_\mathrm{new}$. In detail, our model consists of a feature extractor~$\phi$ and a classifier~$w$. The feature extractor takes an image and outputs a convolutional feature map for prediction,~$\phi^{t}: x \mapsto f^{t}$, where we denote by~$f^{t}(\mathbf{p}) \in \mathbb{R}^D$ a~$D$-dimensional feature vector at position~$\mathbf{p}$. A logit value for a category~$c$ is then computed by the dot product between a feature and a classifier weight for the category~$c$,~$w^{t}_c \in \mathbb{R}^{D}$, as follows:
\begin{equation}
	z^{t}_c(\mathbf{p}) = w^{t}_c \cdot f^{t}(\mathbf{p}),
\end{equation}
where we omit bias terms of the classifier for brevity.

\subsection{Step 1} \label{sec:3.2}
In the first step, we initialize network weights of a model at the stage~$t$ with those for the previous stage~$t-1$, and train a feature extractor and a classifier of the current model with a corresponding dataset~$D^{t}$. A simple way to mitigate catastrophic forgetting for ISS is employing cross-entropy~(CE) and KD~\cite{hinton2014distilling} terms. To this end, we compute probabilities for a category~$c$ at position~$\mathbf{p}$ as follows:
\begin{equation}
	p^{t}_c(\mathbf{p}) = \frac{e^{z^{t}_c(\mathbf{p})}}{\sum_{k \in C^{t}_\mathrm{all}} e^{z^{t}_k(\mathbf{p})}} ,~~~c \in C^{t}_\mathrm{all}
	~~~\text{and}~~~
	q^{t}_c(\mathbf{p}) = \frac{e^{z^{t}_c(\mathbf{p})}}{\sum_{k \in C^{t}_\mathrm{prev}} e^{z^{t}_k(\mathbf{p})}} ,~~~c \in C^{t}_\mathrm{prev},
\end{equation}
which are obtained by applying the softmax function to logit values across~$C^{t}_\mathrm{all}$ and~$C^{t}_\mathrm{prev}$, respectively. The CE and KD terms for ISS are then defined as follows:
\begin{equation} \label{eq:ce_and_kd}
	L_\mathrm{CE}(\mathbf{p})
	= -\log p^{t}_{c^*}(\mathbf{p}),~~~\mathbf{p} \in \mathcal{R}^{t}_\mathrm{new}
	~~~\text{and}~~~
	L_\mathrm{KD}(\mathbf{p})
	= \sum_{k \in C^{t}_\mathrm{prev}}- p^{t-1}_k(\mathbf{p}) \log q^{t}_k(\mathbf{p}),~~~\forall \mathbf{p},
\end{equation}
where~$c^*=y(\mathbf{p})$. The CE term is defined for the labeled regions~$\mathcal{R}^{t}_\mathrm{new}$ only, indicating that logit values of previous categories always decrease. Exploiting the CE term alone is thus prone to overfitting to new categories and leads to catastrophic forgetting. The KD term addresses this problem by transferring the discriminative power of a previous model, trained to classify previous categories, into the current one. Note that $C^{t}_\mathrm{prev} = C^{t-1}_\mathrm{all}$, and the KD term computes softmax probabilities across previous categories without considering new ones. Instead of exploiting CE and KD terms, the seminal work~\cite{MIB20cer} proposes to use CCE and CKD, widely adopted in ISS~\cite{MIB20cer,PLOP21dou,SDR21mic}. In the following, we analyze gradients of CCE and CKD to better understand the influences on ISS, and introduce ALI to train our model.
\begin{wraptable}{r}{0.35\linewidth}
	\aboverulesep=0ex
	\belowrulesep=0ex
	\centering
	\small
	\vspace{.4cm}
	\caption{Gradients of CCE w.r.t~$z^{t}_c$. Note that the optimization process is carried out by gradient descent.}
	\label{tab:grad_cce} 
	\resizebox{\linewidth}{!}
	{
		\begin{tabular}{c !{\vrule width 0.1mm} c !{\vrule width 0.4mm} c}
			\toprule[1pt]
			\multicolumn{2}{c !{\vrule width 0.4mm}}{Conditions} & Gradients
			\\
			\cmidrule[1pt]{1-3}
			\multirow{2}*{$\mathbf{p} \in \mathcal{R}^{t}_\mathrm{new}$} 
			&~$c=y(\mathbf{p})$ 	&~$p^{t}_c - 1$ \\
			\cmidrule{2-3} 
			&~$c\neq y(\mathbf{p})$ &~$p^{t}_c$
			\\
			\cmidrule[1pt]{1-3}
			\multirow{2}*{$\mathbf{p} \notin \mathcal{R}^{t}_\mathrm{new}$}
			&~$c \in C^{t}_\mathrm{new}$ &~$p^{t}_c$ \\
			\cmidrule{2-3} 
			&~$c \in C^{t}_\mathrm{prev}$ &~$p^{t}_c - q^{t}_c$
			\\
			\bottomrule[1pt]
		\end{tabular}
	}
	\vspace{-1.3cm}
\end{wraptable}
\paragraph{CCE.}
The CCE loss~\cite{MIB20cer} additionally computes a probability for unlabeled regions by summing the probabilities over all previous categories as follows:
\begin{equation} \label{eq:grad_cce}
	L_\mathrm{CCE}(\mathbf{p}) = 
	\begin{cases}
		\begin{aligned}
			-\log p^{t}_{c^*}(\mathbf{p}) &,~~~\mathbf{p} \in \mathcal{R}^{t}_\mathrm{new} 
			\\
			-\log p^{t}_\mathrm{cce}(\mathbf{p}) &,~~~\mathbf{p} \notin \mathcal{R}^{t}_\mathrm{new} 
		\end{aligned}
	\end{cases},
\end{equation}
where~$p^{t}_\mathrm{cce}(\mathbf{p}) = \sum_{k \in C^{t}_\mathrm{prev}} p^{t}_k(\mathbf{p})$. This marginal difference from the vanilla CE term brings a significant improvement on ISS. To analyze the reason behind the improvement, we summarize in Table~\ref{tab:grad_cce} gradients of~$L_\mathrm{CCE}$ w.r.t~$z^{t}_c$. The first two rows show that CCE updates logit values for the labeled regions in the same way as in~$L_\mathrm{CE}$. We can see from last two rows that CCE also provides gradients for the unlabeled regions. Specifically, the third row shows that CCE reduces a logit value for new categories by the corresponding probability~$p^{t}_c$, if~$\mathbf{p} \notin \mathcal{R}^{t}_\mathrm{new}$. This is reasonable, since all features on those regions do not at least belong to new categories. CCE thus helps to avoid the overfitting problem, alleviating catastrophic forgetting. Considering the fact that $q^{t}_c$ is always larger than~$p^{t}_c$~(See the supplementary material), we can see from the last row that CCE raises logit values for all previous categories by~$q^{t}_c - p^{t}_c$ in the unlabeled regions. This prevents the logit values of previous categories from continuing to decrease in the unlabeled regions, which is however effective only when predictions of the current model are correct. Otherwise, when predictions are incorrect, CCE rather raises logit values of the wrongly predicted categories, aggravating the incorrect predictions.

\paragraph{CKD.}
Assuming that categories of~$C^{t}_\mathrm{new}$ are likely to be labeled as a background one in the previous stage, the CKD loss~\cite{MIB20cer} is defined as follows:
\begin{align} \label{eq:ckd}
	L_\mathrm{CKD}(\mathbf{p}) 
	= -p^{t-1}_{bg}(\mathbf{p}) \log p^{t}_\mathrm{ckd}(\mathbf{p}) 
	+ \sum_{k \in C^{t}_\mathrm{prev} \backslash \{bg\} } -p^{t-1}_k(\mathbf{p}) \log p^{t}_k(\mathbf{p}) ,~~~\forall\mathbf{p},
\end{align}
where~$p^{t}_\mathrm{ckd}(\mathbf{p}) = \sum_{k \in \{bg\} \cup C^{t}_\mathrm{new}} p^{t}_k(\mathbf{p})$. For previous categories except the background, this term enables transferring knowledge from~$p^{t-1}_k$ to~$p^{t}_k$ directly. Similarly, a vanilla KD term in~Eq.~\eqref{eq:ce_and_kd} also encourages a current model to output probabilities similar to~$p^{t-1}_k$ for~$k \in C^{t}_\mathrm{prev}$, but it distills the knowledge from~$p^{t-1}_k$ into~$q^{t}_k$ without considering logit values for new categories. To compare KD and CKD, we compute gradients w.r.t~$z^{t}_c$ for~$c \in C^{t}_\mathrm{prev} \backslash \{bg\}$ as follows:
\begin{equation} \label{eq:grad_kd_and_ckd}
	\frac{\partial L_\mathrm{KD}(\mathbf{p})}{\partial z^{t}_c(\mathbf{p})}
	= 
	q^{t}_c(\mathbf{p}) - p^{t-1}_c(\mathbf{p})
	~~~\text{and}~~~
	\frac{\partial L_\mathrm{CKD}(\mathbf{p})}{\partial z^{t}_c(\mathbf{p})}
	= 
	p^{t}_c(\mathbf{p}) - p^{t-1}_c(\mathbf{p}).
\end{equation}
Here we provide two cases to explain the reason why CKD better transfers the knowledge for previous categories except the background:~(1) When~$q^{t}_c$ and~$p^{t-1}_c$ are equal, the gradient of KD w.r.t~$z^{t}_c$ is zero, indicating that~$z^{t}_c$ remains the same. However, note that~$p^{t}_c$ is lower than~$p^{t-1}_c$ in this case, since~$p^{t}_c$ is always lower than~$q^{t}_c$. This suggests that~$z^{t}_c$ should rather increase in order that~$p^{t}_c$ follows the target probability~$p^{t-1}_c$. The gradient of CKD w.r.t~$z^{t}_c$ has a negative value in the same case~(\emph{i.e.},~$q^{t}_c=p^{t-1}_c$), suggesting that~$z^{t}_c$ increases by gradient descent.~(2) When the current model already provides the same probability as the target one~(\emph{i.e.},~$p^{t}_c = p^{t-1}_c$), it is unnecessary to adjust~$z^{t}_c$. The gradient of KD w.r.t~$z^{t}_c$ has a positive value, as~$q^{t}_c$ is larger than~$p^{t}_c$~(or equivalently~$p^{t-1}_c$ in this case). This indicates that $z^{t}_c$ decreases by gradient descent, and $p^{t}_c$ in turn becomes lower than~$p^{t-1}_c$. On the other hand, CKD maintains~$z^{t}_c$ the same, since its gradient w.r.t~$z^{t}_c$ is zero. These examples describe the effectiveness of CKD. On the contrary, CKD also has negative effects on ISS. Let us suppose the gradients of CKD w.r.t~$z^{t}_c$ for~$c \in \{bg\} \cup C^{t}_\mathrm{new}$ as follows:
\begin{equation} \label{eq:neg_ckd}
	\frac{\partial L_\mathrm{CKD}(\mathbf{p})}{\partial z^{t}_c(\mathbf{p})}
	=
	\left( p^{t}_\mathrm{ckd}(\mathbf{p}) - p^{t-1}_{bg}(\mathbf{p}) \right) \frac{p^{t}_c(\mathbf{p})}{p^{t}_\mathrm{ckd}(\mathbf{p})}.
\end{equation}

The problem of CKD lies in how it transfers the knowledge from~$p^{t-1}_{bg}$ into~$p^{t}_{bg}$. In particular, CKD makes~$p^{t}_\mathrm{ckd}$ to imitate~$p^{t-1}_{bg}$, instead of directly distilling from~$p^{t-1}_{bg}$ to~$p^{t}_{bg}$. For example, when~$p^{t}_\mathrm{ckd}$ is lower than~$p^{t-1}_{bg}$, the gradient of CKD in~Eq.~\eqref{eq:neg_ckd} has a negative value. Thus, logit values for background and new categories increase by gradient descent. This in turn raises probabilities for background and new categories all together, disturbing discriminating the new categories from the background at training time. In case of~$p^{t}_\mathrm{ckd} > p^{t-1}_{bg}$, CKD reduces the logit values for background and new categories, which is however problematic when $p^{t}_{bg} < p^{t-1}_{bg}$. Note that the logit value for the background category~$z^{t}_{bg}$ should rather increase in this case.
\begin{wraptable}{r}{0.35\linewidth}
	\aboverulesep=0ex
	\belowrulesep=0ex
	\centering
	\small
	\vspace{.35cm}
	\caption{Gradients of ALI w.r.t~$z^{t}_c$.}
	\label{tab:grad_ali}
	\resizebox{\linewidth}{!}
	{
		\begin{tabular}{c !{\vrule width 0.1mm} c !{\vrule width 0.4mm} c}
			\toprule[1pt]
			\multicolumn{2}{c !{\vrule width 0.4mm}}{Conditions} & Gradients
			\\
			\cmidrule[1pt]{1-3}
			\multirow{2}*{$\mathbf{p} \notin \mathcal{R}^{t}_\mathrm{new}$} 
			&~$c \in C^{t}_\mathrm{new}$ &~$p^{t}_c$ \\
			\cmidrule{2-3} 
			&~$c \in C^{t}_\mathrm{prev}$ &~$p^{t}_c-p^{t-1}_c$
			\\
			\bottomrule[1pt]
		\end{tabular}
	}
	\vspace{-1cm}
\end{wraptable}
\paragraph{ALI.}
We have shown that CCE alleviates catastrophic forgetting, while CKD better guides transferring the knowledge of a previous model than KD. To avoid the negative effects of CCE and CKD, we present in Table~\ref{tab:grad_ali} a new form of gradients w.r.t~$z^{t}_c$. The first row is identical to the third one of Table~\ref{tab:grad_cce} that alleviates catastrophic forgetting. The second row is similar to the gradients of CKD in~Eq.~\eqref{eq:grad_kd_and_ckd} that better capture knowledge for previous categories. Note that CKD excludes the background category in~Eq.~\eqref{eq:grad_kd_and_ckd}, since it assumes that new categories of the current stage~($C^{t}_\mathrm{new}$) are marked as the background one at the previous stage. Differently, ours enables computing gradients for all previous categories including the background one~(See the second row), since it does not require the assumption. Accordingly, this form of gradients incorporates the advantages of CCE and CKD, while discarding the negative effects. {Integrating the gradients in Table~\ref{tab:grad_ali} w.r.t~$z^{t}_c$, we define ALI as follows:
\begin{equation} \label{eq:ali}
	L_\mathrm{ALI}(\mathbf{p}) = \log \left(\sum_{k \in C^{t}_\mathrm{all}} e^{z^{t}_k(\mathbf{p})} \right) - \sum_{k \in C^{t}_\mathrm{prev}} p^{t-1}_k(\mathbf{p}) z^{t}_k(\mathbf{p}),~~~\mathbf{p} \notin \mathcal{R}^{t}_\mathrm{new}.
\end{equation}
 The first term applies the log-sum-exp function to logit values across~$C^{t}_\mathrm{all}$, approximating the maximum logit value over~$C^{t}_\mathrm{all}$. The second term computes a weighted average of logit values for~$C^{t}_\mathrm{prev}$ with probabilities of a previous model. In this context, ALI can be viewed as minimizing the difference between the maximum logit value and the weighted average adaptively. That is, it reduces the maximum logit value, while raising the logit values for previous categories. Note that ALI does not require computing either~$p^{t}_\mathrm{cce}$ in CCE or~$p^{t}_\mathrm{ckd}$ in CKD~\cite{MIB20cer}.

\paragraph{Training.}
To train our model, we use CE and ALI terms for labeled and unlabeled regions, respectively. We also apply a vanilla KD term for labeled regions to further regularize our model. An overall objective for the first step is defined as follows: 
\begin{equation} \label{eq:s1}
	L_\mathrm{S1}(\mathbf{p}) 
	= L_\mathrm{CE}(\mathbf{p})
	+ \lambda_\mathrm{ALI} L_\mathrm{ALI}(\mathbf{p})
	+ \lambda_\mathrm{KD} L_\mathrm{KD}(\mathbf{p})\mathbbm{1}[\mathbf{p} \in \mathcal{R}^{t}_\mathrm{new}],
\end{equation}
where$\lambda_\mathrm{ALI}$ and $\lambda_\mathrm{KD}$ are balance parameters. We denote by~$\mathbbm{1}[\cdot]$ an indicator function whose output is 1 if the argument is true, and 0 otherwise.

\subsection{Step 2} \label{sec:3.3}
After training our model, we first extract features of new categories in order to replay them in subsequent stages. Then, we compensate a distribution shift of memorized features, which are extracted in the previous stage~$t-1$, before using them to fine-tune a classifier in the third step. To be specific, we exploit category-specific rotation matrices to update memorized features of each category separately. In the following, we provide detailed descriptions for memorizing features and training matrices. Note that we freeze both a feature extractor~$\phi^{t}$ and a classifier~$w^{t}$ for the second step.

\paragraph{Memorizing features.}
We extract features for prediction~(\emph{i.e.},~$f^t$) and store them to replay in subsequent stages. Specifically, given input images containing one of new categories, a feature extractor first produces feature maps. For each image, features for each category are then averaged using a ground-truth mask. This process is repeated until the number of features for each category reaches a preset number~$S$. That is, we memorize~$S$ features for each category. We provide the pseudo code in the supplementary material.

\paragraph{Training matrices.}
Memorized features, extracted in the previous stage~$t-1$, are not compatible with a current classifier~$w^{t}$. To address this, the work of~\cite{FAN20isc} in image classification proposes to exploit two-layer perceptrons, called FAN, where the number of parameters is roughly $32D^2$. Two main limitations of FAN are as follows:~(1) It updates all features without considering categories. Adopting separate FANs for each category alleviates this issue, but it is computationally expensive.~(2) FAN ignores the relations between memorized features. The structural information in the feature space is crucial for the generalization ability of classifiers. To address these problems, we propose to train category-specific rotation matrices to update features of each category separately. The rotation transform enables maintaining the relations between features within the same category, while the number of parameters for each matrix is $0.5(D^2-D)$ only~(See the supplementary material). To this end, we first define a skew-symmetric matrix~$\textbf{S}_c$ of size~$D \times D$ for the category~$c \in C^{t}_\mathrm{prev}$. A rotation matrix for the category~$c$ is then defined using the Cayley transform as follows:
\begin{equation}
	\textbf{R}_c = (\textbf{I}-\textbf{S}_c)(\textbf{I}+\textbf{S}_c)^{-1},
\end{equation}
where~$\textbf{I}$ indicates an identity matrix. Learning the parameters for the matrices is challenging, since training samples of~$D^{t}$ are labeled only for~$C^{t}_\mathrm{new}$. To handle this, we extract features~$f^{t-1}$ from a previous feature extractor~$\phi^{t-1}$, and compute a correlation score as follows:
\begin{equation}
	v_c(\mathbf{p}) = \sum^{S}_{s=1} \text{ReLU}\left( \frac{f^{t-1}(\mathbf{p})}{\Vert f^{t-1}(\mathbf{p}) \Vert} \cdot \frac{m_c(s)}{\Vert m_c(s) \Vert} \right),
\end{equation}
where we denote by~$m_c(s) \in \mathbb{R}^D$ the~$s$-th item in memorized features of the category~$c$. This allows to identify features associated with previous categories~$C^{t}_\mathrm{prev}$, since $f^{t-1}$ and $m_c(s)$ share the same feature space. We then apply the softmax function over the correlation score as follows:
\begin{equation}
	\sigma_c(\mathbf{p}) = \frac{e^{\tau v_c(\mathbf{p})}}{\sum_\mathbf{p} e^{\tau v_c(\mathbf{p})}},
\end{equation}
where~$\tau$ is a temperature controlling the sharpness of~$\sigma_c$. Note that both features~$f^{t-1}$ and~$f^{t}$ at position~$\mathbf{p}$ encode the same semantic information. Using this fact, we define prototypes of the category~$c$ for previous and current stages, $r^{t-1}_c$ and $r^{t}_c$, respectively, by computing a weighted average of each feature map, as follows:
\begin{equation} \label{eq:prototypes}
	r^{t-1}_c = \sum_\mathbf{p} \sigma_c(\mathbf{p}) f^{t-1}(\mathbf{p}),~~~r^{t}_c = \sum_\mathbf{p} \sigma_c(\mathbf{p}) f^{t}(\mathbf{p}).
\end{equation}
We can match the prototypes for each category~$c$ via the corresponding rotation matrix~$\textbf{R}_c$. Namely, each matrix~$\textbf{R}_c$ rotates a previous prototype~$r^{t-1}_c$ to align it with a current one~$r^{t}_c$. To train the matrix~$\textbf{R}_c$, we define an objective function as follows:
\begin{equation}
	L_\mathrm{S2} = \lambda_\mathrm{ROT}L_\mathrm{FID} + (1-\lambda_\mathrm{ROT})L_\mathrm{REG},
\end{equation}
where we denote by~$L_\mathrm{FID}$ and~$L_\mathrm{REG}$ fidelity and regularization terms, respectively, balanced by the parameter~$\lambda_\mathrm{ROT}$. The fidelity term maximizes cosine similarity as follows:
\begin{equation} \label{eq:fidelity}
	L_\mathrm{FID} = \sum_{c \in C^{t}_\mathrm{prev}} \left( 1-\frac{\hat{r}_c}{\Vert \hat{r}_c \Vert}\cdot\frac{r^{t}_c}{\Vert r^{t}_c \Vert} \right),
\end{equation}
where~$\hat{r}_c = \textbf{R}_c r^{t-1}_c$. This encourages the matrices~$\textbf{R}_c$ to align~$\hat{r}_c$ with~$r^{t}_c$. The regularization term enforces the rotated prototypes~$\hat{r}_c$ to be compatible with the current classifier~$w^{t}$ to better guide the alignment process. To this end, we compute a CE loss using the softmax classifier~$w^{t}$ as follows:
\begin{equation}
	L_\mathrm{REG} = \sum_{c \in C^{t}_\mathrm{prev}} -\log \left( \frac{e^{\hat{r}_c \cdot w^{t}_c}}{\sum_{i \in C^{t}_\mathrm{all}} e^{\hat{r}_c \cdot w^{t}_i}} \right).
\end{equation}

\subsection{Step 3} \label{sec:3.4}
In the third step, we first update memorized features of each category as follows:
\begin{equation}
	\hat{m}_c(s) = \mathbf{R}_c m_c(s).
\end{equation}
The updated features along with training samples of~$D^{t}$ are then used to fine-tune a classifier~$w^{t}$ with the following objective:
\begin{equation} \label{eq:s2}
	L_\mathrm{S3}(\mathbf{p}) 
	= L_\mathrm{FL}(\mathbf{p})
	+ \lambda_\mathrm{ALI} L_\mathrm{ALI}(\mathbf{p})
	+ \lambda_\mathrm{MEM} L_\mathrm{MEM}.
\end{equation}
We denote by~$L_\mathrm{FL}$ and~$L_\mathrm{MEM}$ focal loss~(FL)~\cite{lin2017focal} and CE terms, respectively, defined as follows: 
\begin{equation}
	L_\mathrm{FL}(\mathbf{p})
	=
	-(1-p^{t}_{\hat{c}}(\mathbf{p}))^\alpha \log p^{t}_{\hat{c}}(\mathbf{p}),~~~ 
	\hat{c} = 
	\begin{cases}
		\begin{aligned}
			y(\mathbf{p})~~~~~~~~~~~~~~&,~~~\mathbf{p} \in \mathcal{R}^{t}_\mathrm{new} 
			\\
			\text{argmax}_{k \in C^{t}_\mathrm{prev}} p^{t-1}_k(\mathbf{p}) &,~~~\mathbf{p} \notin \mathcal{R}^{t}_\mathrm{new} 
		\end{aligned}
	\end{cases},
\end{equation}
and
\begin{equation}
	L_\mathrm{MEM}
	= \sum_{c \in C^{t}_\mathrm{prev}} \sum^{S}_{s=1} -\log \left( \frac{e^{\hat{m}_c(s) \cdot w^{t}_c}}{\sum_{k\in C^{t}_\mathrm{all}} e^{\hat{m}_k(s) \cdot w^{t}_k}} \right),
\end{equation}
where we set~$\alpha$ to 2 by default. Following~\cite{SSUL21cha,PLOP21dou,RECALL21mar,SDR21mic}, we mark unlabeled regions in the training samples of~$D^{t}$ as predictions obtained from a previous model on-the-fly. Note that we freeze a feature extractor and the rotation matrices for fine-tuning the classifier.

\section{Experiments}
\vspace{-.2cm}
In this section, we present a quantitative comparison between our method and the state of the art, and show ablation studies. More results including qualitative comparisons can be found in the supplementary material. 
\subsection{Implementation details} \label{sec:4.1}
\paragraph{Dataset and evaluation.}
We evaluate our method on standard ISS benchmarks~(PASCAL VOC~\cite{VOC10ever} and ADE20K~\cite{ADE17zhou}). PASCAL VOC provides~$10,582$ training~\cite{hariharan2011semantic} and~$1,449$ validation samples with 20 object and one background categories, while ADE20K consists of~$20,210$ and~$2,000$ samples for training and validation, respectively, with 150 object/stuff categories. There are three incremental scenarios for each dataset, where we denote by each scenario~$A$-$B$($C$). $A,B$ and~$C$ indicate the number of categories at a base stage, the total number of novel categories, and the number of incremental stages, respectively. We follow the same scenarios as in~\cite{MIB20cer,SSUL21cha,PLOP21dou,RECALL21mar,SDR21mic}, unless otherwise specified. Following the common practice~\cite{SSUL21cha,PLOP21dou}, we focus on an overlapped setting, where unlabeled regions could contain either previous or future categories. For evaluation, we report IoU scores on the validation set for each dataset, and do not exploit test-time augmentation or dense CRF techniques~\cite{krahenbuhl2011efficient}. We denote by mIoU$_\mathrm{base}$, mIoU$_\mathrm{new}$, and mIoU the mean IoU scores over base, new, and all categories, respectively. Computing an IoU score over all categories~(\emph{i.e.} mIoU) is typical, but we have found that this does not reflect IoU scores for new categories well. To address this, as in the evaluation protocol of zero-shot learning methods (\emph{e.g.},~\cite{xian2017zero}), we propose to use the harmonic mean~(hIoU) of mIoU$_\mathrm{base}$ and mIoU$_\mathrm{new}$. For all experiments, we report scores averaged over 3 runs~(\emph{i.e.}, different random seeds).

\paragraph{Training.}
Following~\cite{MIB20cer,SSUL21cha,PLOP21dou,RECALL21mar,SDR21mic}, we adopt DeepLab-V3~\cite{chen2017rethinking} with ResNet-101~\cite{res16he}. ResNet-101 is initialized with pre-trained weights for classification on ImageNet~\cite{deng2009imagenet}. We use the SGD optimizer with an initial learning rate set to~1e-2 and~1e-3 for base and incremental stages, respectively. DeepLab-V3 is trained for 30 and 60 epochs at a base stage~($t=1$) on PASCAL VOC~\cite{VOC10ever} and ADE20K~\cite{ADE17zhou}, respectively. For each incremental stage~($t>1$), we perform a cross-validation to choose the number of epochs on PASCAL VOC, while fixing it to 60 on ADE20K. We train rotation matrices for 10 epochs using the Adam optimizer with an initial learning rate of~1e-3, and fix a preset number~$S$ and a temperature value~$\tau$ to~$1,000$ and 10 for all experiments. We fine-tune a classifier for 1 epoch using the SGD optimizer with an initial learning rate of~1e-3. For all experiments, we adjust the learning rate by the poly schedule. We provide a detailed description of hyperparameter settings in the supplementary material.

\begin{table}[t!]
  \setlength{\tabcolsep}{0.21em}
  \centering
  \scriptsize
  \caption{Quantitative results on ADE20K~\cite{ADE17zhou} in terms of IoU scores. SSUL-M~\cite{SSUL21cha} uses a replay buffer that consists of 300 previously seen images together with corresponding ground-truth labels. Numbers in bold are the best performance, while underlined ones are the second best. We show standard deviations in parentheses. Numbers for other methods~\cite{MIB20cer,ILT19umb} are taken from SSUL.~$\dagger$: Results are obtained by running the source codes provided by the authors.}
  \label{tab:sota_ade}
  \begin{tabular}{L{1.43cm} C{0.98cm}C{0.93cm}C{0.8cm}C{0.8cm}
  						    C{0.98cm}C{0.93cm}C{0.8cm}C{0.8cm}
  						    C{0.98cm}C{0.93cm}C{0.8cm}C{0.8cm}}
      \toprule
      \multirow{2.5}{*}{Methods}
      & \multicolumn{4}{c}{\textbf{100-50(1)}}
      & \multicolumn{4}{c}{\textbf{50-100(2)}}
      & \multicolumn{4}{c}{\textbf{100-50(5)}} 
      \\
      
      \cmidrule(lr){2-5} \cmidrule(lr){6-9} \cmidrule(lr){10-13}
      & \tiny{mIoU$_\mathrm{base}$} & \tiny{mIoU$_\mathrm{new}$} & mIoU & hIoU
      & \tiny{mIoU$_\mathrm{base}$} & \tiny{mIoU$_\mathrm{new}$} & mIoU & hIoU
      & \tiny{mIoU$_\mathrm{base}$} & \tiny{mIoU$_\mathrm{new}$} & mIoU & hIoU 
      \\
      
      \midrule
      \midrule
  	  \multicolumn{13}{c}{Without memorized images or features}
      \\
      
      \midrule
      \midrule
      ILT~\cite{ILT19umb}
      & 18.29 & 14.40 & 17.00 & 16.11
	  &~~3.53 & 12.85 &~~9.70 &~~5.54
      &~~0.08 &~~1.31 &~~0.49 &~~0.15
      \\

      MiB~\cite{MIB20cer}
      & 40.52 & 17.17 & 32.79 & 24.12 
	  & 45.57 & 21.01 & 29.31 & 28.76
	  & 38.21 & 11.12 & 29.24 & 17.23
      \\

      \multirow{1.7}{*}{PLOP$^\dagger$~\cite{PLOP21dou}}
      & \underline{42.10} & {16.22} & {33.53} & {23.42}
      & {48.24} & \underline{21.31} & \underline{30.40} & \underline{29.56}
      & \underline{40.78} & {14.02} & {31.92} & {20.87} 
      \\ [-0.25em]
      & {\fontsize{6}{0}\selectfont(0.02)}
      & {\fontsize{6}{0}\selectfont(0.15)}
      & {\fontsize{6}{0}\selectfont(0.06)}
      & {\fontsize{6}{0}\selectfont(0.15)}
      & {\fontsize{6}{0}\selectfont(0.03)}
      & {\fontsize{6}{0}\selectfont(0.08)}
      & {\fontsize{6}{0}\selectfont(0.06)}
      & {\fontsize{6}{0}\selectfont(0.08)}
      & {\fontsize{6}{0}\selectfont(0.04)}
      & {\fontsize{6}{0}\selectfont(0.03)}
      & {\fontsize{6}{0}\selectfont(0.04)}
      & {\fontsize{6}{0}\selectfont(0.03)}
      \\
          
	  SSUL~\cite{SSUL21cha}
      & {41.28} & \underline{18.02} & \underline{33.58} & \underline{25.09}
      & \underline{48.38} & {20.15} & {29.56} & {28.45}
      & {40.20} & \underline{18.75} & \underline{33.10} & \underline{25.57}
      \\
      
	  \multirow{1.5}{*}{ALIFE}
      & \textbf{42.18} & \textbf{23.07} & \textbf{35.86} & \textbf{29.83}
      & \textbf{48.98} & \textbf{25.69} & \textbf{33.56} & \textbf{33.70} 
      & \textbf{41.02} & \textbf{22.76} & \textbf{34.98} & \textbf{29.28} 
      \\ [-0.25em]
      & {\fontsize{6}{0}\selectfont(0.08)}
      & {\fontsize{6}{0}\selectfont(0.51)}
      & {\fontsize{6}{0}\selectfont(0.12)}
      & {\fontsize{6}{0}\selectfont(0.41)}
      & {\fontsize{6}{0}\selectfont(0.12)}
      & {\fontsize{6}{0}\selectfont(0.20)}
      & {\fontsize{6}{0}\selectfont(0.16)}
      & {\fontsize{6}{0}\selectfont(0.19)}
      & {\fontsize{6}{0}\selectfont(0.23)}
      & {\fontsize{6}{0}\selectfont(0.55)}
      & {\fontsize{6}{0}\selectfont(0.34)}
      & {\fontsize{6}{0}\selectfont(0.51)}
      \\

      \midrule
      \midrule
  	  \multicolumn{13}{c}{With memorized images or features}
  	  \\

      \midrule
      \midrule
      SSUL-M~\cite{SSUL21cha}
      & \textbf{42.79} & \underline{17.54} & \underline{34.37} & \underline{24.88}
      & \textbf{49.12} & \underline{20.10} & \underline{29.77} & \underline{28.53}
      & \textbf{42.86} & \underline{17.66} & \underline{34.46} & \underline{25.01}
      \\

      \cline{1-13}
      &&&&&&&&&&&
      \\ [-0.8em]

	  \multirow{1.75}{*}{ALIFE-M}
      & \underline{42.28} & \textbf{23.58} & \textbf{36.09} & \textbf{30.28}
      & \underline{48.99} & \textbf{26.15} & \textbf{33.87} & \textbf{34.10}
      & \underline{41.17} & \textbf{23.07} & \textbf{35.18} & \textbf{29.57}
      \\[-0.25em]
      & {\fontsize{6}{0}\selectfont(0.05)}
      & {\fontsize{6}{0}\selectfont(0.45)}
      & {\fontsize{6}{0}\selectfont(0.12)}
      & {\fontsize{6}{0}\selectfont(0.36)}
      & {\fontsize{6}{0}\selectfont(0.07)}
      & {\fontsize{6}{0}\selectfont(0.10)}
      & {\fontsize{6}{0}\selectfont(0.09)}
      & {\fontsize{6}{0}\selectfont(0.10)}
      & {\fontsize{6}{0}\selectfont(0.21)}
      & {\fontsize{6}{0}\selectfont(0.22)}
      & {\fontsize{6}{0}\selectfont(0.21)}
      & {\fontsize{6}{0}\selectfont(0.23)}
      \\

      \bottomrule
  \end{tabular}
  \vspace{-.5cm}
\end{table}

\subsection{Comparison with the state of the art}
\paragraph{ADE20K.}
We compare in Table~\ref{tab:sota_ade} our approach with state-of-the-art methods, including MiB~\cite{MIB20cer}, PLOP~\cite{PLOP21dou}, and SSUL~\cite{SSUL21cha}. Note that RECALL~\cite{RECALL21mar} is not designed to handle stuff categories, and results on ADE20K~\cite{ADE17zhou} are not available. From this table, we have three findings as follows:~(1) Our approach exploiting the first step only, denoted by ALIFE, already outperforms all other methods in terms of both mIoU and hIoU scores by significant margins for all scenarios. This validates the effectiveness of our approach without memorizing features. In particular, we can see that ALIFE even outperforms SSUL-M~\cite{SSUL21cha} that memorizes~$300$ images along with ground-truth labels for replaying. A plausible reason is that SSUL freezes a feature extractor, limiting the flexibility to deal with new categories.~(2)~ALIFE shows substantial IoU gains over MiB~\cite{MIB20cer} using CCE and CKD for all scenarios. This verifies that both CCE and CKD are not always helpful for ISS. ALI is free from the limitations of CCE and CKD, and it allows our model to better learn new categories without forgetting previous ones.~(3) Our approach memorizing features, denoted by ALIFE-M, improves the performance over ALIFE in terms of all metrics for all scenarios. Note that SSUL-M even performs worse than SSUL for 100-50(1) and 100-50(6) cases. Considering that we rely on at least~$9$ times less memory requirements than SSUL-M for 100-50(1) and 50-100(2) cases, the gains from memorizing features are remarkable compared to those of SSUL-M over SSUL.

\paragraph{PASCAL VOC.}
We show in Table~\ref{tab:sota_voc} quantitative results on PASCAL VOC~\cite{VOC10ever}. Note that a comparison of SSUL~\cite{SSUL21cha}~(SSUL-M) and other methods including ours is not fair, since it additionally exploits an off-the-shelf saliency detector~\cite{hou2017deeply} that brings significant improvements on PASCAL VOC. From this table, we can see that ALIFE outperforms all other methods in terms of hIoU scores for 20-1(1) and 16-5(1) scenarios, further demonstrating the effectiveness of our approach without memorizing features. We can also see that ALIFE-M gives substantial gains over ALIFE in terms of all metrics for all scenarios. In particular, the largest mIoU and hIoU gains of 2.18\% and 3.60\%, respectively, are reported on 16-5(5). This confirms once again that our feature replay scheme is effective even for the most challenging scenario on PASCAL VOC. SSUL-M~\cite{SSUL21cha} and RECALL~\cite{RECALL21mar} largely outperform ours only for 16-5(5), but they require at least 604 and 15 times more memory footprint than ALIFE-M, respectively. Moreover, RECALL has difficulty for handling stuff categories, and SSUL performs poorly on ADE20K~\cite{ADE17zhou}, where the saliency detector is not applicable. Our approach is versatile in that it is free to handle stuff categories without performance degradation on ADE20K~(See Table~\ref{tab:sota_ade}).

\begin{table}[t!]
  \setlength{\tabcolsep}{0.21em}
  \centering
  \scriptsize
  \caption{Quantitative results on PASCAL VOC~\cite{VOC10ever} in terms of IoU scores. SSUL-M~\cite{SSUL21cha} memorizes 100 images in total, while RECALL~\cite{RECALL21mar} uses 500 images for each previous category. Note that both SSUL and SSUL-M also require an off-the-shelf saliency detector~\cite{hou2017deeply} on PASCAL VOC. Numbers in bold are the best performance, while underlined ones are the second best. We show standard deviations in parentheses. Numbers for other methods are taken from corresponding papers.~$\dagger$: Results are obtained with the source codes provided by the authors.}
  \label{tab:sota_voc}
  \begin{tabular}{L{1.43cm} C{0.98cm}C{0.93cm}C{0.8cm}C{0.8cm}
  						    C{0.98cm}C{0.93cm}C{0.8cm}C{0.8cm}
  						    C{0.98cm}C{0.93cm}C{0.8cm}C{0.8cm}}
      \toprule
      \multirow{2.5}{*}{Methods}
      & \multicolumn{4}{c}{\textbf{20-1(1)}}
      & \multicolumn{4}{c}{\textbf{16-5(1)}}
      & \multicolumn{4}{c}{\textbf{16-5(5)}} 
      \\

      \cmidrule(lr){2-5} \cmidrule(lr){6-9} \cmidrule(lr){10-13}
      & \tiny{mIoU$_\mathrm{base}$} & \tiny{mIoU$_\mathrm{new}$} & mIoU & hIoU
      & \tiny{mIoU$_\mathrm{base}$} & \tiny{mIoU$_\mathrm{new}$} & mIoU & hIoU
      & \tiny{mIoU$_\mathrm{base}$} & \tiny{mIoU$_\mathrm{new}$} & mIoU & hIoU 
      \\
      
      \midrule
      \midrule
  	  \multicolumn{13}{c}{Without memorized images or features}
      \\

      \midrule
      \midrule
	  EWC~\cite{kirkpatrick2017overcoming}
      & 26.90 & 14.00 & 26.30 & 18.42
      & 24.30 & 35.50 & 27.10 & 28.85
      &~~0.30 &~~4.30 &~~1.30 &~~0.56
      \\ 

	  LwF-MC~\cite{li2017learning}
      & 64.40 & 13.30 & 61.90 & 22.05
      & 58.10 & 35.00 & 52.30 & 43.68
      &~~6.40 &~~8.40 &~~6.90 &~~7.26
      \\ 
      
	  ILT~\cite{ILT19umb}
      & 67.75 & 10.88 & 65.05 & 18.75
      & 67.08 & 39.23 & 60.45 & 49.51
      &~~8.75 &~~7.99 &~~8.56 &~~8.35 
      \\ 
      
      \multirow{1.5}{*}{MiB$^\dagger$~\cite{MIB20cer}}
      & 70.42 & 17.70 & 67.91 & 28.25
      & 76.68 & 49.03 & 70.09 & 59.81
      & 37.98 & 12.28 & 31.86 & 18.56 
      \\ [-0.25em]
      & {\fontsize{6}{0}\selectfont(0.13)}
      & {\fontsize{6}{0}\selectfont(1.89)}
      & {\fontsize{6}{0}\selectfont(0.17)}
      & {\fontsize{6}{0}\selectfont(2.44)}
      & {\fontsize{6}{0}\selectfont(0.11)}
      & {\fontsize{6}{0}\selectfont(0.27)}
      & {\fontsize{6}{0}\selectfont(0.14)}
      & {\fontsize{6}{0}\selectfont(0.23)}
      & {\fontsize{6}{0}\selectfont(0.72)}
      & {\fontsize{6}{0}\selectfont(0.19)}
      & {\fontsize{6}{0}\selectfont(0.57)}
      & {\fontsize{6}{0}\selectfont(0.28)}
      \\ 
      
      SDR~\cite{SDR21mic}
      & 71.30 & 23.40 & 69.00 & 35.24 
      & 76.30 & \underline{50.20} & 70.10 & 60.56
      & 47.30 & 14.70 & 39.50 & 22.43 
      \\ 

      \multirow{1.7}{*}{PLOP$^\dagger$~\cite{PLOP21dou}}
      & {75.89} & \underline{34.90} & {73.94} & \underline{47.81}
      & 76.37 & 49.55 & 69.98 & 60.10
      & \underline{64.51} & {19.93} & {53.90} & {30.45} 
      \\ [-0.25em]
      & {\fontsize{6}{0}\selectfont(0.21)}
      & {\fontsize{6}{0}\selectfont(0.93)}
      & {\fontsize{6}{0}\selectfont(0.24)}
      & {\fontsize{6}{0}\selectfont(0.91)}
      & {\fontsize{6}{0}\selectfont(0.14)}
      & {\fontsize{6}{0}\selectfont(0.29)}
      & {\fontsize{6}{0}\selectfont(0.18)}
      & {\fontsize{6}{0}\selectfont(0.26)}
      & {\fontsize{6}{0}\selectfont(0.13)}
      & {\fontsize{6}{0}\selectfont(0.20)}
      & {\fontsize{6}{0}\selectfont(0.06)}
      & {\fontsize{6}{0}\selectfont(0.21)}
      \\ 

	  SSUL~\cite{SSUL21cha}
      & \textbf{77.73} & 29.68 & \textbf{75.44} & 42.96
      & \textbf{77.82} & 50.10 & \underline{71.22} & \underline{60.96}
      & \textbf{77.31} & \textbf{36.59} & \textbf{67.61} & \textbf{49.67}
      \\ 
         
      \multirow{1.5}{*}{ALIFE}
      & \underline{76.61} & \textbf{49.36} & \underline{75.31} & \textbf{60.03}
      & \underline{77.18} & \textbf{52.52} & \textbf{71.31} & \textbf{62.50} 
      & {64.44} & \underline{34.91} & \underline{57.41} & \underline{45.29} 
      \\ [-0.25em]
      & {\fontsize{6}{0}\selectfont(0.52)}
      & {\fontsize{6}{0}\selectfont(1.01)}
      & {\fontsize{6}{0}\selectfont(0.52)}
      & {\fontsize{6}{0}\selectfont(0.84)}
      & {\fontsize{6}{0}\selectfont(0.66)}
      & {\fontsize{6}{0}\selectfont(0.48)}
      & {\fontsize{6}{0}\selectfont(0.52)}
      & {\fontsize{6}{0}\selectfont(0.41)}
      & {\fontsize{6}{0}\selectfont(1.24)}
      & {\fontsize{6}{0}\selectfont(1.05)}
      & {\fontsize{6}{0}\selectfont(1.19)}
      & {\fontsize{6}{0}\selectfont(1.18)}
      \\

      \midrule
      \midrule
  	  \multicolumn{13}{c}{With memorized images or features}
      \\

      \midrule
      \midrule
      SSUL-M~\cite{SSUL21cha}
      & \textbf{78.83} & {49.76} & \textbf{76.49} & {61.01}
      & \textbf{78.40} & \textbf{55.80} & \textbf{73.02} & \textbf{65.20}
      & \textbf{78.36} & \underline{49.01} & \textbf{71.37} & \textbf{60.30}
      \\

	  RECALL~\cite{RECALL21mar}
      & {68.10} & \textbf{55.30} & {68.60} & \underline{61.04}
      & {67.70} & {54.30} & {65.60} & {60.26} 
      & \underline{67.80} & \textbf{50.90} & \underline{64.80} & \underline{58.15} 
      \\

	  \multirow{1.75}{*}{ALIFE-M}
      & \underline{76.72} & \underline{52.29} & \underline{75.56} & \textbf{62.19}
      & \underline{77.66} & \underline{55.27} & \underline{72.33} & \underline{64.57}
      & {66.09} & {38.81} & {59.59} & {48.89}
      \\[-0.25em]
      & {\fontsize{6}{0}\selectfont(0.57)}
      & {\fontsize{6}{0}\selectfont(0.62)}
      & {\fontsize{6}{0}\selectfont(0.55)}
      & {\fontsize{6}{0}\selectfont(0.49)}
      & {\fontsize{6}{0}\selectfont(0.36)}
      & {\fontsize{6}{0}\selectfont(0.98)}
      & {\fontsize{6}{0}\selectfont(0.21)}
      & {\fontsize{6}{0}\selectfont(0.59)}
      & {\fontsize{6}{0}\selectfont(0.64)}
      & {\fontsize{6}{0}\selectfont(1.86)}
      & {\fontsize{6}{0}\selectfont(0.93)}
      & {\fontsize{6}{0}\selectfont(1.64)}
      \\
      
      \bottomrule
  \end{tabular}
  \vspace{-.5cm}
\end{table}

\subsection{Discussion}
\vspace{-.15cm}
\paragraph{The ordering of categories.}
\begin{wrapfigure}{r}{0.45\linewidth}
	\vspace{-1.2cm}
	\centering
	\includegraphics[width=\linewidth]{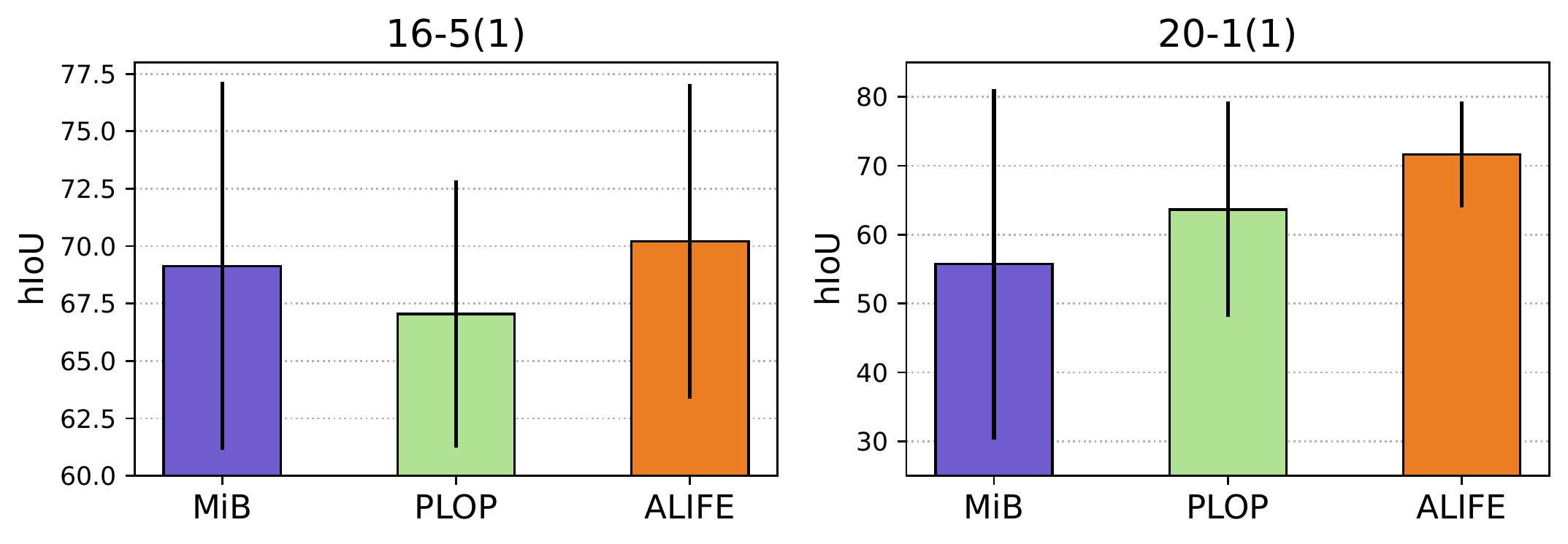}
	\vspace{-.6cm}
	\caption{Comparison of average hIoU scores over three different category orderings on PASCAL VOC~\cite{VOC10ever}.}
	\label{fig:diff_orders}
	\vspace{-.6cm}
\end{wrapfigure}
To show the robustness of our approach to different category orderings, we generate two random category orderings for 16-5(1) and 20-1(1) cases, and report the mean and standard deviation of hIoU scores over three orderings: one alphabetical and two random orderings. We compare in Fig.~\ref{fig:diff_orders} our approach without memorizing features, denoted by ALIFE, to MiB~\cite{MIB20cer} and PLOP~\cite{PLOP21dou}. Results of MiB and PLOP are obtained by running the source codes provided by the authors. We can clearly see that ALIFE produces better results than other methods on both 16-5(1) and 20-1(1) cases.

\paragraph{Ablation study on the first step.}
\begin{wraptable}{r}{0.45\linewidth}
	\centering
	\small
	\vspace{-.45cm}
	\caption{Comparison of IoU scores using different loss terms of our approach on 16-5(1) of PASCAL VOC~\cite{VOC10ever}. \textit{Labeled}, \textit{Unlabeled}, and \textit{All} indicate that KD is applied for labeled, unlabeled, and all regions, respectively.}
	\label{tab:abla_first}
	\vspace{-.25cm}
	\resizebox{\linewidth}{!}
	{
		\begin{tabular}{c c ccc cccc}
			\toprule[1pt]
			\multirow{2.5}{*}{CE} & \multirow{2.5}{*}{ALI} & \multicolumn{3}{c}{KD} & \multirow{2.5}{*}{mIoU$_\mathrm{base}$} & \multirow{2.5}{*}{mIoU$_\mathrm{new}$} & \multirow{2.5}{*}{mIoU} & \multirow{2.5}{*}{hIoU}
			\\
			
			\cmidrule(lr){3-5}
            & & \textit{Labeled}&\textit{Unlabeled}&\textit{All} & & &
			\\
			
			\cmidrule[1pt]{1-9}
			\checkmark & & & & & 16.45 &~~6.94 & 14.19 &~~9.74
			\\
			
			\checkmark & \checkmark & & & & 75.50 & 49.81 & 69.39 & 60.02
			\\
			
			\checkmark & \checkmark & \checkmark & & & \textbf{77.18} & \textbf{52.52} & \textbf{71.31} & \textbf{62.50}
			\\
			
			\checkmark & \checkmark & & \checkmark & & 75.45 & 49.75 & 69.33 & 59.97
			\\
			
			\checkmark & \checkmark & & & \checkmark & \underline{76.25} & \underline{50.99} & \underline{70.24} & \underline{61.12}
			\\
			
			\bottomrule[1pt]
		\end{tabular}
	}
	\vspace{-.2cm}
\end{wraptable}
We show in Table~\ref{tab:abla_first} an ablation analysis on different loss terms of our approach without memorizing features~(Eq.~\eqref{eq:s1}). The first row shows that using the CE term alone performs poorly due to catastrophic forgetting. We can see from the second row that ALI alleviates the catastrophic forgetting problem remarkably. The third row shows that additionally using the KD term for labeled regions on top of CE and ALI terms further gives IoU gains. On the other hand, we can see from the fourth row that applying the KD term for unlabeled regions rather degrades the performance. Since our ALI already encourages a current model to imitate knowledge of a previous one for unlabeled regions, we conjecture that additionally applying the KD term for those regions is redundant. The last row shows that using the KD term for all regions is beneficial to improving the performance slightly.

\paragraph{Ablation study on the third step.}
\begin{wraptable}{r}{0.45\linewidth}
	\centering
	\small
	\vspace{-.45cm}
	\caption{Comparison of IoU scores using different loss terms of our approach on 16-5(1) of PASCAL VOC~\cite{VOC10ever}. \textit{Labeled}: CE or FL is applied only for labeled regions. \textit{All$^*$}: To apply CE or FL for all regions, we mark unlabeled regions as predictions of a previous model on-the-fly.}
	\label{tab:abla_third}
	\vspace{-.25cm}
	\resizebox{\linewidth}{!}
	{
		\begin{tabular}{cc cc cc C{1.25cm}C{1.25cm}C{1.25cm}C{1.25cm}}
				\toprule[1pt]
				\multicolumn{2}{c}{CE} & \multicolumn{2}{c}{FL} & \multirow{2.5}{*}{ALI} & \multirow{2.5}{*}{MEM} & \multirow{2.5}{*}{mIoU$_\mathrm{base}$} & \multirow{2.5}{*}{mIoU$_\mathrm{new}$} & \multirow{2.5}{*}{mIoU} & \multirow{2.5}{*}{hIoU}
				\\
				\cmidrule(lr){1-2} \cmidrule(lr){3-4}
				\textit{Labeled}&\textit{All$^*$}& \textit{Labeled}&\textit{All$^*$} & & & & & &
				\\
				\cmidrule[1pt]{1-10}
				
				\checkmark & & & & & & 76.11 & 48.03 & 69.42 & 58.89
				\\
				
				& \checkmark & & & & & 76.71 & 50.37 & 70.44 & 60.81
				\\
				
				& & \checkmark & & & & 76.51 & 49.70 & 70.13 & 60.26	
				\\
				
	            & & & \checkmark & & & 76.81 & 50.98 & 70.66 & 61.29	
				\\
				
	            & & & \checkmark & & \checkmark & \underline{77.24} & \underline{54.90} & \underline{71.92} & \underline{64.17}	
				\\
				
	            & & & \checkmark & \checkmark & & 77.25 & 52.88 & 71.44 & 62.78	
	            \\
	            
	            & & & \checkmark & \checkmark & \checkmark & \textbf{77.66} & \textbf{55.27} & \textbf{72.33} & \textbf{64.57}	
	            \\
	
				\bottomrule[1pt]
		\end{tabular}
	}
	\vspace{-.4cm}
\end{wraptable}
We report in Table~\ref{tab:abla_third} IoU scores for different losses of our method in the third step~(Eq.~\eqref{eq:s2}). From the first four rows, we can see that 1) using CE or FL terms alone degrade the performance. This is because training samples of a current dataset mainly contain new categories, causing the class imbalance between previous and new categories; 2) FL provides better results than CE; 3) The pseudo labeling strategy works favorably for both CE and FL, mitigating the imbalance. From the last three rows, we can see that 1) replaying 1K features for each previous category lessens the influence of the class imbalance problem significantly and 2) ALI is also helpful to alleviate the imbalance problem, further boosting the performance.

\paragraph{Limitation.}
Our approach to memorizing features for an experience replay reduces memory requirements significantly and avoids data privacy issues. Nonetheless, it requires more memory than other methods~\cite{MIB20cer,PLOP21dou,SDR21mic} that do not adopt the experience replay. Since our approach to using ALI without memorizing features shows state-of-the-art results on standard ISS benchmarks~\cite{VOC10ever,ADE17zhou}, we believe that ALI could give useful insights for developing ISS methods that do not rely on the replay strategy. It is also worth noting that all existing methods including ours assume that newly incoming samples are clean and reliable. However, in practice, training samples of a new task might be biased and unreliable. This raises new concerns:~1) addressing noisy samples during training and~2) memorizing clean samples only. Handling these potential risks would also be an interesting future direction for ISS.

\vspace{-.2cm}
\section{Conclusion}
\vspace{-.2cm}
We have introduced a new ISS method, ALIFE, that alleviates catastrophic forgetting and reduces memory requirements for an experience replay. First, we have presented a gradient analysis of CCE and CKD for better understanding the effects on catastrophic forgetting, and have proposed ALI incorporating the merits of CCE and CKD. Second, we have proposed a feature replay scheme using the Cayley transform that requires less memory footprint than memorizing raw images. Finally, we have shown that ALIFE sets a new state of the art on standard ISS benchmarks.

\vspace{-.2cm}
\begin{ack}
\vspace{-.2cm}
This work was supported by Institute of Information \& Communications Technology Planing \& Evaluation (IITP) grant funded by the Korea government (MSIT) (No.RS-2022-00143524, Development of Fundamental Technology and Integrated Solution for Next-Generation Automatic Artificial Intelligence System) and the Yonsei Signature Research Cluster Program of 2022 (2022-22-0002).
\end{ack}

\bibliographystyle{plainnat}
\bibliography{reference}

\begin{thebibliography}{40}
\providecommand{\natexlab}[1]{#1}
\providecommand{\url}[1]{\texttt{#1}}
\expandafter\ifx\csname urlstyle\endcsname\relax
  \providecommand{\doi}[1]{doi: #1}\else
  \providecommand{\doi}{doi: \begingroup \urlstyle{rm}\Url}\fi

\bibitem[Castro et~al.(2018)Castro, Mar{\'\i}n-Jim{\'e}nez, Guil, Schmid, and
  Alahari]{castro2018end}
Francisco~M Castro, Manuel~J Mar{\'\i}n-Jim{\'e}nez, Nicol{\'a}s Guil, Cordelia
  Schmid, and Karteek Alahari.
\newblock End-to-end incremental learning.
\newblock In \emph{ECCV}, 2018.

\bibitem[Cermelli et~al.(2020)Cermelli, Mancini, Bulo, Ricci, and
  Caputo]{MIB20cer}
Fabio Cermelli, Massimiliano Mancini, Samuel~Rota Bulo, Elisa Ricci, and
  Barbara Caputo.
\newblock Modeling the background for incremental learning in semantic
  segmentation.
\newblock In \emph{CVPR}, 2020.

\bibitem[Cha et~al.(2021)Cha, Yoo, Moon, et~al.]{SSUL21cha}
Sungmin Cha, YoungJoon Yoo, Taesup Moon, et~al.
\newblock S{SUL}: Semantic segmentation with unknown label for exemplar-based
  class-incremental learning.
\newblock \emph{NeurIPS}, 2021.

\bibitem[Chaudhry et~al.(2018)Chaudhry, Dokania, Ajanthan, and
  Torr]{chaudhry2018riemannian}
Arslan Chaudhry, Puneet~K Dokania, Thalaiyasingam Ajanthan, and Philip~HS Torr.
\newblock Riemannian walk for incremental learning: Understanding forgetting
  and intransigence.
\newblock In \emph{ECCV}, 2018.

\bibitem[Chen et~al.(2017)Chen, Papandreou, Schroff, and
  Adam]{chen2017rethinking}
Liang-Chieh Chen, George Papandreou, Florian Schroff, and Hartwig Adam.
\newblock Rethinking atrous convolution for semantic image segmentation.
\newblock \emph{arXiv}, 2017.

\bibitem[Chopra et~al.(2005)Chopra, Hadsell, and LeCun]{chopra2005learning}
Sumit Chopra, Raia Hadsell, and Yann LeCun.
\newblock Learning a similarity metric discriminatively, with application to
  face verification.
\newblock In \emph{CVPR}, 2005.

\bibitem[Deng et~al.(2009)Deng, Dong, Socher, Li, Li, and
  Fei-Fei]{deng2009imagenet}
Jia Deng, Wei Dong, Richard Socher, Li-Jia Li, Kai Li, and Li~Fei-Fei.
\newblock Image{N}et: A large-scale hierarchical image database.
\newblock In \emph{CVPR}, 2009.

\bibitem[Douillard et~al.(2020)Douillard, Cord, Ollion, Robert, and
  Valle]{POD20dou}
Arthur Douillard, Matthieu Cord, Charles Ollion, Thomas Robert, and Eduardo
  Valle.
\newblock P{ODN}et: Pooled outputs distillation for small-tasks incremental
  learning.
\newblock In \emph{ECCV}, 2020.

\bibitem[Douillard et~al.(2021)Douillard, Chen, Dapogny, and Cord]{PLOP21dou}
Arthur Douillard, Yifu Chen, Arnaud Dapogny, and Matthieu Cord.
\newblock P{LOP}: Learning without forgetting for continual semantic
  segmentation.
\newblock In \emph{CVPR}, 2021.

\bibitem[Everingham et~al.(2010)Everingham, Van~Gool, Williams, Winn, and
  Zisserman]{VOC10ever}
Mark Everingham, Luc Van~Gool, Christopher~KI Williams, John Winn, and Andrew
  Zisserman.
\newblock The {PASCAL} visual object classes ({VOC}) challenge.
\newblock \emph{IJCV}, 88\penalty0 (2):\penalty0 303--338, 2010.

\bibitem[French(1999)]{french1999catastrophic}
Robert~M French.
\newblock Catastrophic forgetting in connectionist networks.
\newblock \emph{Trends in cognitive sciences}, 3\penalty0 (4):\penalty0
  128--135, 1999.

\bibitem[Goodfellow et~al.(2014)Goodfellow, Pouget-Abadie, Mirza, Xu,
  Warde-Farley, Ozair, Courville, and Bengio]{goodfellow2014generative}
Ian Goodfellow, Jean Pouget-Abadie, Mehdi Mirza, Bing Xu, David Warde-Farley,
  Sherjil Ozair, Aaron Courville, and Yoshua Bengio.
\newblock Generative adversarial nets.
\newblock \emph{NeurIPS}, 2014.

\bibitem[Hariharan et~al.(2011)Hariharan, Arbel{\'a}ez, Bourdev, Maji, and
  Malik]{hariharan2011semantic}
Bharath Hariharan, Pablo Arbel{\'a}ez, Lubomir Bourdev, Subhransu Maji, and
  Jitendra Malik.
\newblock Semantic contours from inverse detectors.
\newblock In \emph{ICCV}, 2011.

\bibitem[He et~al.(2016)He, Zhang, Ren, and Sun]{res16he}
Kaiming He, Xiangyu Zhang, Shaoqing Ren, and Jian Sun.
\newblock Deep residual learning for image recognition.
\newblock In \emph{CVPR}, 2016.

\bibitem[Hinton et~al.(2014)Hinton, Vinyals, and Dean]{hinton2014distilling}
Geoffrey Hinton, Oriol Vinyals, and Jeff Dean.
\newblock Distilling the knowledge in a neural network.
\newblock In \emph{NeurIPS Workshop}, 2014.

\bibitem[Hou et~al.(2017)Hou, Cheng, Hu, Borji, Tu, and Torr]{hou2017deeply}
Qibin Hou, Ming-Ming Cheng, Xiaowei Hu, Ali Borji, Zhuowen Tu, and Philip~HS
  Torr.
\newblock Deeply supervised salient object detection with short connections.
\newblock In \emph{CVPR}, 2017.

\bibitem[Hou et~al.(2020)Hou, Zhang, Cheng, and Feng]{hou2020strip}
Qibin Hou, Li~Zhang, Ming-Ming Cheng, and Jiashi Feng.
\newblock Strip {P}ooling: Rethinking spatial pooling for scene parsing.
\newblock In \emph{CVPR}, 2020.

\bibitem[Hou et~al.(2019)Hou, Pan, Loy, Wang, and Lin]{hou2019learning}
Saihui Hou, Xinyu Pan, Chen~Change Loy, Zilei Wang, and Dahua Lin.
\newblock Learning a unified classifier incrementally via rebalancing.
\newblock In \emph{CVPR}, 2019.

\bibitem[Iscen et~al.(2020)Iscen, Zhang, Lazebnik, and Schmid]{FAN20isc}
Ahmet Iscen, Jeffrey Zhang, Svetlana Lazebnik, and Cordelia Schmid.
\newblock Memory-efficient incremental learning through feature adaptation.
\newblock In \emph{ECCV}, 2020.

\bibitem[Kemker and Kanan(2018)]{kemker2017fearnet}
Ronald Kemker and Christopher Kanan.
\newblock Fear{N}et: Brain-inspired model for incremental learning.
\newblock \emph{ICLR}, 2018.

\bibitem[Kirkpatrick et~al.(2017)Kirkpatrick, Pascanu, Rabinowitz, Veness,
  Desjardins, Rusu, Milan, Quan, Ramalho, Grabska-Barwinska,
  et~al.]{kirkpatrick2017overcoming}
James Kirkpatrick, Razvan Pascanu, Neil Rabinowitz, Joel Veness, Guillaume
  Desjardins, Andrei~A Rusu, Kieran Milan, John Quan, Tiago Ramalho, Agnieszka
  Grabska-Barwinska, et~al.
\newblock Overcoming catastrophic forgetting in neural networks.
\newblock \emph{Proceedings of the national academy of sciences}, 114\penalty0
  (13):\penalty0 3521--3526, 2017.

\bibitem[Kr{\"a}henb{\"u}hl and Koltun(2011)]{krahenbuhl2011efficient}
Philipp Kr{\"a}henb{\"u}hl and Vladlen Koltun.
\newblock Efficient inference in fully connected crfs with gaussian edge
  potentials.
\newblock \emph{NeurIPS}, 2011.

\bibitem[Li and Hoiem(2017)]{li2017learning}
Zhizhong Li and Derek Hoiem.
\newblock Learning without forgetting.
\newblock \emph{IEEE TPAMI}, 40\penalty0 (12):\penalty0 2935--2947, 2017.

\bibitem[Lin et~al.(2017)Lin, Goyal, Girshick, He, and
  Doll{\'a}r]{lin2017focal}
Tsung-Yi Lin, Priya Goyal, Ross Girshick, Kaiming He, and Piotr Doll{\'a}r.
\newblock Focal loss for dense object detection.
\newblock In \emph{ICCV}, 2017.

\bibitem[Lopez-Paz and Ranzato(2017)]{lopez2017gradient}
David Lopez-Paz and Marc'Aurelio Ranzato.
\newblock Gradient episodic memory for continual learning.
\newblock \emph{NeurIPS}, 2017.

\bibitem[Maracani et~al.(2021)Maracani, Michieli, Toldo, and
  Zanuttigh]{RECALL21mar}
Andrea Maracani, Umberto Michieli, Marco Toldo, and Pietro Zanuttigh.
\newblock R{ECALL}: Replay-based continual learning in semantic segmentation.
\newblock In \emph{ICCV}, 2021.

\bibitem[McCloskey and Cohen(1989)]{mccloskey1989catastrophic}
Michael McCloskey and Neal~J Cohen.
\newblock Catastrophic interference in connectionist networks: The sequential
  learning problem.
\newblock In \emph{Psychology of learning and motivation}, volume~24, pages
  109--165. Elsevier, 1989.

\bibitem[Mermillod et~al.(2013)Mermillod, Bugaiska, and
  Bonin]{mermillod2013stability}
Martial Mermillod, Aur{\'e}lia Bugaiska, and Patrick Bonin.
\newblock The stability-plasticity dilemma: Investigating the continuum from
  catastrophic forgetting to age-limited learning effects.
\newblock \emph{Frontiers in psychology}, 4:\penalty0 504, 2013.

\bibitem[Michieli and Zanuttigh(2019)]{ILT19umb}
Umberto Michieli and Pietro Zanuttigh.
\newblock Incremental learning techniques for semantic segmentation.
\newblock In \emph{ICCV Workshop}, 2019.

\bibitem[Michieli and Zanuttigh(2021)]{SDR21mic}
Umberto Michieli and Pietro Zanuttigh.
\newblock Continual semantic segmentation via repulsion-attraction of sparse
  and disentangled latent representations.
\newblock In \emph{CVPR}, 2021.

\bibitem[Ostapenko et~al.(2019)Ostapenko, Puscas, Klein, Jahnichen, and
  Nabi]{ostapenko2019learning}
Oleksiy Ostapenko, Mihai Puscas, Tassilo Klein, Patrick Jahnichen, and Moin
  Nabi.
\newblock Learning to remember: A synaptic plasticity driven framework for
  continual learning.
\newblock In \emph{CVPR}, 2019.

\bibitem[Prabhu et~al.(2020)Prabhu, Torr, and Dokania]{prabhu2020gdumb}
Ameya Prabhu, Philip~HS Torr, and Puneet~K Dokania.
\newblock G{D}umb: A simple approach that questions our progress in continual
  learning.
\newblock In \emph{ECCV}, 2020.

\bibitem[Rebuffi et~al.(2017)Rebuffi, Kolesnikov, Sperl, and
  Lampert]{rebuffi2017icarl}
Sylvestre-Alvise Rebuffi, Alexander Kolesnikov, Georg Sperl, and Christoph~H
  Lampert.
\newblock i{C}a{RL}: Incremental classifier and representation learning.
\newblock In \emph{CVPR}, 2017.

\bibitem[Shin et~al.(2017)Shin, Lee, Kim, and Kim]{shin2017continual}
Hanul Shin, Jung~Kwon Lee, Jaehong Kim, and Jiwon Kim.
\newblock Continual learning with deep generative replay.
\newblock \emph{NeurIPS}, 2017.

\bibitem[Shokri and Shmatikov(2015)]{shokri2015privacy}
Reza Shokri and Vitaly Shmatikov.
\newblock Privacy-preserving deep learning.
\newblock In \emph{ACM SIGSAC CCS}, 2015.

\bibitem[Wu et~al.(2019)Wu, Chen, Wang, Ye, Liu, Guo, and Fu]{wu2019large}
Yue Wu, Yinpeng Chen, Lijuan Wang, Yuancheng Ye, Zicheng Liu, Yandong Guo, and
  Yun Fu.
\newblock Large scale incremental learning.
\newblock In \emph{CVPR}, 2019.

\bibitem[Xian et~al.(2017)Xian, Schiele, and Akata]{xian2017zero}
Yongqin Xian, Bernt Schiele, and Zeynep Akata.
\newblock Zero-shot learning-the good, the bad and the ugly.
\newblock In \emph{CVPR}, 2017.

\bibitem[Xiang et~al.(2019)Xiang, Fu, Ji, and Huang]{xiang2019incremental}
Ye~Xiang, Ying Fu, Pan Ji, and Hua Huang.
\newblock Incremental learning using conditional adversarial networks.
\newblock In \emph{ICCV}, 2019.

\bibitem[Zhao et~al.(2020)Zhao, Xiao, Gan, Zhang, and Xia]{zhao2020maintaining}
Bowen Zhao, Xi~Xiao, Guojun Gan, Bin Zhang, and Shu-Tao Xia.
\newblock Maintaining discrimination and fairness in class incremental
  learning.
\newblock In \emph{CVPR}, 2020.

\bibitem[Zhou et~al.(2017)Zhou, Zhao, Puig, Fidler, Barriuso, and
  Torralba]{ADE17zhou}
Bolei Zhou, Hang Zhao, Xavier Puig, Sanja Fidler, Adela Barriuso, and Antonio
  Torralba.
\newblock Scene parsing through {ADE}20{K} dataset.
\newblock In \emph{CVPR}, 2017.

\end{thebibliography}

\section*{Checklist}
\begin{enumerate}

\item For all authors...
\begin{enumerate}
  \item Do the main claims made in the abstract and introduction accurately reflect the paper's contributions and scope?
    \answerYes{}
  \item Did you describe the limitations of your work?
    \answerYes{}
  \item Did you discuss any potential negative societal impacts of your work?
    \answerNA{}
  \item Have you read the ethics review guidelines and ensured that your paper conforms to them?
    \answerYes{}
\end{enumerate}

\item If you are including theoretical results...
\begin{enumerate}
  \item Did you state the full set of assumptions of all theoretical results?
    \answerYes{} We describe in Sec.~\ref{sec:3.2} the assumption of CKD. Our ALI does not require any assumptions.
        \item Did you include complete proofs of all theoretical results?
    \answerYes{We provide detailed derivations in the supplementary material.}
\end{enumerate}

\item If you ran experiments...
\begin{enumerate}
  \item Did you include the code, data, and instructions needed to reproduce the main experimental results (either in the supplemental material or as a URL)?
    \answerYes{}
  \item Did you specify all the training details (e.g., data splits, hyperparameters, how they were chosen)?
    \answerYes{We describe implementation details in Sec.~\ref{sec:4.1}. For more details, please refer to the supplementary material.}
        \item Did you report error bars (e.g., with respect to the random seed after running experiments multiple times)?
    \answerYes{For all experiments, we report averaged scores over 3 runs. For example, we report standard deviations in Tables~\ref{tab:sota_ade} and~\ref{tab:sota_voc}.}
  \item Did you include the total amount of compute and the type of resources used (e.g., type of GPUs, internal cluster, or cloud provider)?
    \answerYes{We use 2 NVIDIA TITAN RTX GPUs for all experiments. Please refer to the supplementary material for details.}
\end{enumerate}

\item If you are using existing assets (e.g., code, data, models) or curating/releasing new assets...
\begin{enumerate}
  \item If your work uses existing assets, did you cite the creators?
    \answerYes{We train DeepLab-V3~\cite{chen2017rethinking} on PASCAL VOC~\cite{VOC10ever} and ADE20K~\cite{VOC10ever}.}
  \item Did you mention the license of the assets?
    \answerNA{}
  \item Did you include any new assets either in the supplemental material or as a URL?
    \answerNA{}
  \item Did you discuss whether and how consent was obtained from people whose data you're using/curating?
    \answerNA{}
  \item Did you discuss whether the data you are using/curating contains personally identifiable information or offensive content?
    \answerNA{}
\end{enumerate}

\item If you used crowdsourcing or conducted research with human subjects...
\begin{enumerate}
  \item Did you include the full text of instructions given to participants and screenshots, if applicable?
    \answerNA{}
  \item Did you describe any potential participant risks, with links to Institutional Review Board (IRB) approvals, if applicable?
    \answerNA{}
  \item Did you include the estimated hourly wage paid to participants and the total amount spent on participant compensation?
    \answerNA{}
\end{enumerate}

\end{enumerate}



\clearpage


\section*{Supplementary material}

# ALIFE: Adaptive Logit Regularizer and Feature Replay for Incremental Semantic Segmentation Supplement

**Youngmin Oh**   **Donghyeon Baek**   **Bumsub Ham***
School of Electrical and Electronic Engineering, Yonsei University
https://cvlab.yonsei.ac.kr/projects/ALIFE

Here we show detailed derivations together with a further analysis of CCE and CKD (Sec. A). We describe more details for experimental settings and hyperparamters, and present a pseudo code of our approach (Sec. B). We also provide more discussions on design choices and a feature replay scheme (Sec. C), and show more quantitative and qualitative results (Sec. D).

## A  Derivations together with an analysis of CCE and CKD

**Proposition 1.** *For $c \in C^t_{\text{prev}}$, $q^t_c$ is always larger than $p^t_c$.*

*Proof.*

$$\begin{aligned}
\frac{q^t_c(\mathbf{p})}{p^t_c(\mathbf{p})} &= \frac{e^{z^t_c(\mathbf{p})}}{\sum_{i \in C^t_{\text{prev}}} e^{z^t_i(\mathbf{p})}} \frac{\sum_{j \in C^t_{\text{all}}} e^{z^t_j(\mathbf{p})}}{e^{z^t_c(\mathbf{p})}} \\
&= \frac{\sum_{j \in C^t_{\text{all}}} e^{z^t_j(\mathbf{p})}}{\sum_{i \in C^t_{\text{prev}}} e^{z^t_i(\mathbf{p})}} \\
&= \frac{\sum_{i \in C^t_{\text{prev}}} e^{z^t_i(\mathbf{p})} + \sum_{j \in C^t_{\text{new}}} e^{z^t_j(\mathbf{p})}}{\sum_{i \in C^t_{\text{prev}}} e^{z^t_i(\mathbf{p})}} \qquad \text{(i)} \\
&= 1 + \frac{\sum_{j \in C^t_{\text{new}}} e^{z^t_j(\mathbf{p})}}{\sum_{i \in C^t_{\text{prev}}} e^{z^t_i(\mathbf{p})}} \\
&= 1 + \gamma(\mathbf{p}).
\end{aligned}$$

Since $\gamma(\mathbf{p})$ is always larger than zero, we can derive the following inequality:

$$\frac{q^t_c(\mathbf{p})}{p^t_c(\mathbf{p})} = 1 + \gamma(\mathbf{p}) > 1. \qquad \text{(ii)}$$

□

---

*Corresponding author.

36th Conference on Neural Information Processing Systems (NeurIPS 2022).

## A.1 CCE

We first reformulate the CCE loss (Eq. (5) in the main paper) as follows:

$$
L_{\text{CCE}}(\mathbf{p}) = \begin{cases} -\log p^t_{c^*}(\mathbf{p}), & \mathbf{p} \in \mathcal{R}^t_{\text{new}} \\ -\log p^t_{\text{cce}}(\mathbf{p}), & \mathbf{p} \notin \mathcal{R}^t_{\text{new}} \end{cases}
$$

$$
= \begin{cases} -z^t_{c^*}(\mathbf{p}) + \log\left(\sum_{k \in C^t_{\text{all}}} e^{z^t_k(\mathbf{p})}\right), & \mathbf{p} \in \mathcal{R}^t_{\text{new}} \\ -\log\left(\sum_{i \in C^t_{\text{prev}}} e^{z^t_i(\mathbf{p})}\right) + \log\left(\sum_{k \in C^t_{\text{all}}} e^{z^t_k(\mathbf{p})}\right), & \mathbf{p} \notin \mathcal{R}^t_{\text{new}} \end{cases}. \quad \text{(iii)}
$$

Then, the gradient of CCE w.r.t $z^t_c$ is computed as follows:

$$
\frac{\partial L_{\text{CCE}}(\mathbf{p})}{\partial z^t_c(\mathbf{p})} = \begin{cases} -\mathbb{1}[c = y(\mathbf{p})] + \dfrac{e^{z^t_c(\mathbf{p})}}{\sum_{k \in C^t_{\text{all}}} e^{z^t_k(\mathbf{p})}}, & \mathbf{p} \in \mathcal{R}^t_{\text{new}} \\ -\dfrac{e^{z^t_c(\mathbf{p})}}{\sum_{i \in C^t_{\text{prev}}} e^{z^t_i(\mathbf{p})}} \mathbb{1}[c \in C^t_{\text{prev}}] + \dfrac{e^{z^t_c(\mathbf{p})}}{\sum_{k \in C^t_{\text{all}}} e^{z^t_k(\mathbf{p})}}, & \mathbf{p} \notin \mathcal{R}^t_{\text{new}} \end{cases}
$$

$$
= \begin{cases} -\mathbb{1}[c = y(\mathbf{p})] + p^t_c(\mathbf{p}), & \mathbf{p} \in \mathcal{R}^t_{\text{new}} \\ -q^t_c(\mathbf{p}) \mathbb{1}[c \in C^t_{\text{prev}}] + p^t_c(\mathbf{p}), & \mathbf{p} \notin \mathcal{R}^t_{\text{new}} \end{cases} \quad \text{(iv)}
$$

$$
= \begin{cases} p^t_c(\mathbf{p}) - 1, & \mathbf{p} \in \mathcal{R}^t_{\text{new}} \text{ and } c = y(\mathbf{p}) \\ p^t_c(\mathbf{p}), & \mathbf{p} \in \mathcal{R}^t_{\text{new}} \text{ and } c \neq y(\mathbf{p}) \\ p^t_c(\mathbf{p}), & \mathbf{p} \notin \mathcal{R}^t_{\text{new}} \text{ and } c \in C^t_{\text{new}} \\ p^t_c(\mathbf{p}) - q^t_c(\mathbf{p}), & \mathbf{p} \notin \mathcal{R}^t_{\text{new}} \text{ and } c \in C^t_{\text{prev}} \end{cases}.
$$

We can see that the gradients of Eq. (iv) are the same as those of Table 1 in the main paper. The last row of Table 1 suggests that logit values for all previous categories always increase by $q^t_c - p^t_c$ in the unlabeled regions. We visualize in Fig. A heatmaps of $q^t_c - p^t_c$ for $c \in \{background, chair, person\}$. For visualization, MiB [2] is trained with samples for a tv category after learning 20 categories (e.g., background, chair, and person) on PASCAL VOC [7]. We can see that CCE raises logit values for chair and person categories, even though corresponding ground-truth labels are likely to be the background category. This in turn lessens the discriminability of the model for the previous categories.

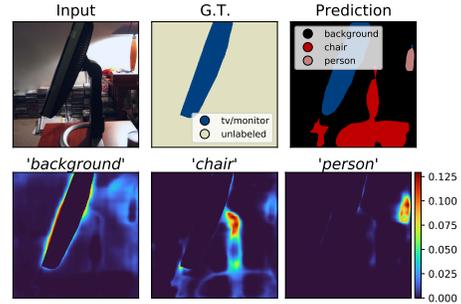

Figure A: The negative effect of CCE on ISS.

## A.2 CKD

The CKD loss (Eq. (6) in the main paper) can be reformulated as follows:

$$L_{\text{CKD}}(\mathbf{p}) = -p_{bg}^{t-1}(\mathbf{p}) \log p_{\text{ckd}}^t(\mathbf{p}) + \sum_{k \in C_{\text{prev}}^t \setminus \{bg\}} -p_k^{t-1}(\mathbf{p}) \log p_k^t(\mathbf{p})$$

$$= -p_{bg}^{t-1}(\mathbf{p}) \log \left( \sum_{i \in \{bg\} \cup C_{\text{new}}^t} e^{z_i^t(\mathbf{p})} \right) + p_{bg}^{t-1}(\mathbf{p}) \log \left( \sum_{j \in C_{\text{all}}^t} e^{z_j^t(\mathbf{p})} \right)$$
$$+ \sum_{k \in C_{\text{prev}}^t \setminus \{bg\}} \left( -p_k^{t-1}(\mathbf{p}) z_k^t(\mathbf{p}) + p_k^{t-1}(\mathbf{p}) \log \left( \sum_{j \in C_{\text{all}}^t} e^{z_j^t(\mathbf{p})} \right) \right)$$

$$= -p_{bg}^{t-1}(\mathbf{p}) \log \left( \sum_{i \in \{bg\} \cup C_{\text{new}}^t} e^{z_i^t(\mathbf{p})} \right) + \left( \sum_{k \in C_{\text{prev}}^t} p_k^{t-1}(\mathbf{p}) \right) \cdot \log \left( \sum_{j \in C_{\text{all}}^t} e^{z_j^t(\mathbf{p})} \right)$$
$$+ \sum_{k \in C_{\text{prev}}^t \setminus \{bg\}} -p_k^{t-1}(\mathbf{p}) z_k^t(\mathbf{p})$$

$$= -p_{bg}^{t-1}(\mathbf{p}) \log \left( \sum_{i \in \{bg\} \cup C_{\text{new}}^t} e^{z_i^t(\mathbf{p})} \right) + \log \left( \sum_{j \in C_{\text{all}}^t} e^{z_j^t(\mathbf{p})} \right) + \sum_{k \in C_{\text{prev}}^t \setminus \{bg\}} -p_k^{t-1}(\mathbf{p}) z_k^t(\mathbf{p}).$$
(v)

Note that $\sum_{k \in C_{\text{prev}}^t} p_k^{t-1}(\mathbf{p}) = 1$. We then compute the gradient of CKD w.r.t $z_c^t$ as follows:

$$\frac{\partial L_{\text{CKD}}(\mathbf{p})}{\partial z_c^t(\mathbf{p})} = -p_{bg}^{t-1}(\mathbf{p}) \frac{e^{z_c^t(\mathbf{p})}}{\sum_{i \in \{bg\} \cup C_{\text{new}}^t} e^{z_i^t(\mathbf{p})}} \mathbb{1}[c \in \{bg\} \cup C_{\text{new}}^t] + \frac{e^{z_c^t(\mathbf{p})}}{\sum_{j \in C_{\text{all}}^t} e^{z_j^t(\mathbf{p})}} - p_c^{t-1}(\mathbf{p}) \mathbb{1}[c \in C_{\text{prev}}^t \setminus \{bg\}]$$

$$= -p_{bg}^{t-1}(\mathbf{p}) \frac{e^{z_c^t(\mathbf{p})}}{\sum_{i \in \{bg\} \cup C_{\text{new}}^t} e^{z_i^t(\mathbf{p})}} \frac{\sum_{k \in C_{\text{all}}^t} e^{z_k^t(\mathbf{p})}}{\sum_{k \in C_{\text{all}}^t} e^{z_k^t(\mathbf{p})}} \mathbb{1}[c \in \{bg\} \cup C_{\text{new}}^t] + p_c^t(\mathbf{p}) - p_c^{t-1}(\mathbf{p}) \mathbb{1}[c \in C_{\text{prev}}^t \setminus \{bg\}]$$

$$= -p_{bg}^{t-1}(\mathbf{p}) \frac{p_c^t(\mathbf{p})}{\sum_{i \in \{bg\} \cup C_{\text{new}}^t} p_i^t(\mathbf{p})} \mathbb{1}[c \in \{bg\} \cup C_{\text{new}}^t] + p_c^t(\mathbf{p}) - p_c^{t-1}(\mathbf{p}) \mathbb{1}[c \in C_{\text{prev}}^t \setminus \{bg\}]$$

$$= -p_{bg}^{t-1}(\mathbf{p}) \frac{p_c^t(\mathbf{p})}{p_{\text{ckd}}^t(\mathbf{p})} \mathbb{1}[c \in \{bg\} \cup C_{\text{new}}^t] + p_c^t(\mathbf{p}) - p_c^{t-1}(\mathbf{p}) \mathbb{1}[c \in C_{\text{prev}}^t \setminus \{bg\}]$$

$$= \begin{cases} p_c^t(\mathbf{p}) - p_c^{t-1}(\mathbf{p}), & c \in C_{\text{prev}}^t \setminus \{bg\} \\ p_c^t(\mathbf{p}) - p_{bg}^{t-1}(\mathbf{p}) \frac{p_c^t(\mathbf{p})}{p_{\text{ckd}}^t(\mathbf{p})}, & c \in \{bg\} \cup C_{\text{new}}^t \end{cases}$$

$$= \begin{cases} p_c^t(\mathbf{p}) - p_c^{t-1}(\mathbf{p}), & c \in C_{\text{prev}}^t \setminus \{bg\} \\ \left(1 - \frac{p_{bg}^{t-1}(\mathbf{p})}{p_{\text{ckd}}^t(\mathbf{p})}\right) p_c^t(\mathbf{p}), & c \in \{bg\} \cup C_{\text{new}}^t \end{cases}$$

$$= \begin{cases} p_c^t(\mathbf{p}) - p_c^{t-1}(\mathbf{p}), & c \in C_{\text{prev}}^t \setminus \{bg\} \\ \left(p_{\text{ckd}}^t(\mathbf{p}) - p_{bg}^{t-1}(\mathbf{p})\right) \frac{p_c^t(\mathbf{p})}{p_{\text{ckd}}^t(\mathbf{p})}, & c \in \{bg\} \cup C_{\text{new}}^t \end{cases}.$$
(vi)

We can see that $p_c^t - p_c^{t-1}$ and $(p_{\text{ckd}}^t - p_{bg}^{t-1}) \frac{p_c^t}{p_{\text{ckd}}^t}$ in Eq. (vi) are the same as the gradients of CKD in the main paper (See Eqs. (7) and (8)).

To further compare KD and CKD, we first assume that a previous model outputs confident predictions. Namely, probabilities of the previous model show one clear peak as follows:

$$p_{\hat{y}}^{t-1}(\mathbf{p}) \approx 1, \tag{vii}$$

where $\hat{y} = \operatorname{argmax}_{k \in C_{\text{prev}}^t} p_k^{t-1}(\mathbf{p})$. We empirically show that this assumption is valid in Fig. B. We then simplify KD and CKD terms (See Eqs. (4) and (6) in the main paper) as follows:

$$L_{\text{KD}}(\mathbf{p}) \approx -p_{\hat{y}}^{t-1}(\mathbf{p}) \log q_{\hat{y}}^t(\mathbf{p}) \quad \text{and} \quad L_{\text{CKD}}(\mathbf{p}) \approx \begin{cases} -p_{bg}^{t-1}(\mathbf{p}) \log p_{\text{ckd}}^t(\mathbf{p}), & \hat{y} \in \{bg\} \\ -p_{\hat{y}}^{t-1}(\mathbf{p}) \log p_{\hat{y}}^t(\mathbf{p}), & \hat{y} \in C_{\text{prev}}^t \setminus \{bg\} \end{cases}. \tag{viii}$$

We compute gradients of each term w.r.t $z_{\hat{y}}^t$ for $\hat{y} \in C_{\text{prev}}^t \setminus \{bg\}$ as follows:

$$\frac{\partial L_{\text{KD}}(\mathbf{p})}{\partial z_{\hat{y}}^t(\mathbf{p})} \approx q_{\hat{y}}^t(\mathbf{p}) - p_{\hat{y}}^{t-1}(\mathbf{p}) = \alpha_{\text{kd}}(\mathbf{p}) \quad \text{and} \quad \frac{\partial L_{\text{CKD}}(\mathbf{p})}{\partial z_{\hat{y}}^t(\mathbf{p})} \approx p_{\hat{y}}^t(\mathbf{p}) - p_{\hat{y}}^{t-1}(\mathbf{p}) = \alpha_{\text{ckd}}(\mathbf{p}). \tag{ix}$$

Table A summarizes the comparison between KD and CKD into three cases. In the case of ①, a current model produces the probability $p_{\hat{y}}^t$ lower than the target one $p_{\hat{y}}^{t-1}$. Thus, $z_{\hat{y}}^t$ should increase in order that $p_{\hat{y}}^t$ follows $p_{\hat{y}}^{t-1}$. Although both KD and CKD raise $z_{\hat{y}}^t$, we can see that CKD raises $z_{\hat{y}}^t$ more strongly than its counterpart ($|\alpha_{\text{kd}}| < |\alpha_{\text{ckd}}|$). This suggests that CKD helps the current model to produce $p_{\hat{y}}^t$ similar to $p_{\hat{y}}^{t-1}$ more quickly.

Table A: Comparison of gradients of KD and CKD w.r.t $z_{\hat{y}}^t$ for $\hat{y} \in C_{\text{prev}}^t \setminus \{bg\}$.

| case ① $(p_{\hat{y}}^{t-1} \geq q_{\hat{y}}^t > p_{\hat{y}}^t)$ | $0 \geq \alpha_{\text{kd}} > \alpha_{\text{ckd}}$ |
|---|---|
| case ② $(q_{\hat{y}}^t > p_{\hat{y}}^{t-1} \geq p_{\hat{y}}^t)$ | $\alpha_{\text{kd}} > 0 \geq \alpha_{\text{ckd}}$ |
| case ③ $(q_{\hat{y}}^t > p_{\hat{y}}^t \geq p_{\hat{y}}^{t-1})$ | $\alpha_{\text{kd}} > \alpha_{\text{ckd}} \geq 0$ |

In the case of ②, the current model outputs $p_{\hat{y}}^t$ lower than $p_{\hat{y}}^{t-1}$, indicating that $z_{\hat{y}}^t$ needs to increase. However, note that the probability $q_{\hat{y}}^t$ computed without considering logit values of new categories is larger than the target one $p_{\hat{y}}^{t-1}$ (*i.e.*, $\alpha_{\text{kd}} > 0$). Hence, KD rather reduces $z_{\hat{y}}^t$, which results in reducing $p_{\hat{y}}^t$. On the other hand, CKD raises $z_{\hat{y}}^t$ in order that $p_{\hat{y}}^t$ reaches $p_{\hat{y}}^{t-1}$. In the case of ③, the current model already produces the probability $p_{\hat{y}}^t$ larger than the target one $p_{\hat{y}}^{t-1}$. In this case, it is unnecessary to reduce the logit value $z_{\hat{y}}^t$, since reducing $z_{\hat{y}}^t$ could make the probability $p_{\hat{y}}^t$ lower than the target one $p_{\hat{y}}^{t-1}$. KD however reduces $z_{\hat{y}}^t$ more strongly than CKD ($|\alpha_{\text{kd}}| > |\alpha_{\text{ckd}}|$).

We plot in Fig. B the ratio of each case during training. To this end, we extract feature maps from a mini-batch and compute the number of features belonging to each case. We denote by $N_1$, $N_2$, and $N_3$ the number of features in each case, respectively. Then, the ratio of each case is defined as follows:

$$\text{ratio}_i = \frac{N_i}{N}, \quad i \in \{1, 2, 3\}, \tag{x}$$

where $N$ indicates the total number of features, *i.e.*, $N = N_1 + N_2 + N_3$. We compute the ra-

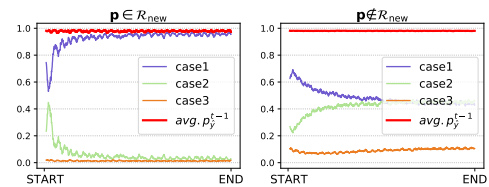

Figure B: Ratio curves of each case in labeled regions (left) and unlabeled ones (right) during training. We also plot the average target probability $p_{\hat{y}}^{t-1}$ (red curves) during training.

tio of each case at every iteration during training. For visualization, we train MiB [2] on 20-1(1) of PASCAL VOC [7]. We have the following observations: (1) We can see that the average target probability $p_{\hat{y}}^{t-1}$ (red curves) is roughly one during training, validating the assumption in Eq. (vii). We can also see that the third case is negligible for both labeled and unlabeled regions; (2) The first case is the most dominant in the labeled regions (See Fig. B left), which is natural in that the current model is likely to output low probabilities $p_{\hat{y}}^t$ for previous categories in those regions[2]. Since CKD raises $z_{\hat{y}}^t$ more strongly than KD in the case of ①, we can interpret that it encourages the current model to imitate the previous one more strongly in the labeled regions. This suggests that

---

[2]Note that the labeled regions contain new categories only.

Table B: Hyperparameter settings. EP: the number of epochs.

| Datasets | Scenarios | $t$ | Step 1 | | | Step 3 | | |
|---|---|---|---|---|---|---|---|---|
| | | | $\lambda_{\text{ALI}}$ | $\lambda_{\text{KD}}$ | EP | $\lambda_{\text{ALI}}$ | $\lambda_{\text{MEM}}$ | EP |
| ADE20K [15] | **100-50(1)** | 2 | 1 | 1 | 60 | 0 | 0.5 | 1 |
| | **50-100(2)** | 2 | 1 | 20 | 60 | 0 | 0.1 | 1 |
| | | 3 | 1 | 20 | 60 | 0 | 0.5 | 1 |
| | **100-50(5)** | 2 | 1 | 1 | 60 | 5 | 2 | 1 |
| | | 3 | 1 | 1 | 60 | 5 | 1 | 1 |
| | | 4 | 1 | 1 | 60 | 2 | 1 | 1 |
| | | 5 | 1 | 1 | 60 | 0 | 0.1 | 1 |
| | | 6 | 1 | 1 | 60 | 0 | 0.5 | 1 |
| PASCAL VOC [7] | **20-1(1)** | 2 | 1 | 1 | 5 | 1 | 1 | 1 |
| | **16-5(1)** | 2 | 2 | 1 | 10 | 1 | 10 | 1 |
| | **16-5(5)** | 2 | 3 | 1 | 10 | 3 | 1 | 1 |
| | | 3 | 5 | 10 | 5 | 5 | 20 | 1 |
| | | 4 | 2 | 1 | 5 | 2 | 1 | 1 |
| | | 5 | 3 | 10 | 5 | 3 | 2 | 1 |
| | | 6 | 2 | 1 | 5 | 1 | 1 | 1 |

CKD acts as a strong regularizer for the current model in those regions; (3) The first two cases are prevalent in the unlabeled regions (See Fig. B right). Note that predictions of the previous model are likely to be correct in unlabeled regions[3]. In this context, it is important for the current model to accurately and quickly imitate such predictions (*i.e.*, $p_{\hat{y}}^{t-1}$) in those regions in order to preserve the knowledge for previous categories. Considering that CKD helps the current model to produce the target probability $p_{\hat{y}}^{t-1}$ more quickly and accurately than KD in the cases of ① and ②, respectively, CKD is more favorable in the unlabeled regions. Note that KD rather prevents the current model from producing the target probability in the case of ②. These empirical studies once again explain the reason why CKD outperforms KD.

## B More details

**Training rotation matrices.** Here we provide a detailed description for training rotation matrices. We first define a strictly upper triangular matrix $\mathbf{U}_c$ of size $D \times D$ for a category $c \in C_{\text{prev}}^t$ as follows:

$$\mathbf{U}_c = \begin{pmatrix} 0 & u_{1,2} & u_{1,3} & \dots & u_{1,D} \\ 0 & 0 & u_{2,3} & \dots & u_{2,D} \\ \vdots & & \ddots & & \vdots \\ 0 & 0 & 0 & \dots & u_{D-1,D} \\ 0 & 0 & 0 & \dots & 0 \end{pmatrix}, \quad \text{(xi)}$$

where we denote by $u_{i,j}$ the element in the $i$-th row and $j$-th column. Note that the number of non-zero elements for the matrix $\mathbf{U}_c$ is $0.5(D^2 - D)$. Then, the skew-symmetric matrix $\mathbf{S}_c$ can be defined using the corresponding upper triangular matrix $\mathbf{U}_c$ as follows:

$$\mathbf{S}_c = \mathbf{U}_c - \mathbf{U}_c^\top. \quad \text{(xii)}$$

The skew-symmetric matrix is used to define the rotation matrix $\mathbf{R}_c$ as in Eq. (11) of the main paper. We train the rotation matrices with the objective function in Eq. (15) (See Sec. 3.3 in the main paper). To be specific, the elements of triangular matrices are trained with random initialization. Thus, the number of parameters for each rotation matrix is $0.5(D^2 - D)$ only.

**Pseudo code of our approach.** We summarize in Algorithm 1 an overall procedure of our approach for incremental stages ($t > 1$). Note that a training process at a base stage ($t = 1$) is equivalent to that of fully-supervised segmentation models.

---

[3]This is because unlabeled regions do not at least contain new categories and the previous model has been trained to classify previous categories.

**Experimental details.** Following the common practice, we have adopted DeepLab-V3 [4] with ResNet-101 [8] pre-trained for ImageNet Classification [5]. In particular, SDR [13] and SSUL [3] use ResNet-101 provided by PyTorch [14], while MiB [2] and PLOP [6] exploit a variant of ResNet-101 using in-place ABN [1] layers. Following SDR and SSUL, we have adopted ResNet-101 provided by PyTorch. For all experiments, we implement our approach using PyTorch, and use two NVIDIA TITAN RTX GPUs along with an Intel i5-9600K CPU. We train DeepLab-V3 and rotation matrices with a batch size of 24, *i.e.*, 12 samples for each GPU.

**Hyperparameter settings.** We empirically set the value of $\tau$ to 10 in order to ensure that correlation scores are sharp enough. For the focal loss [11], we use the default hyperparameters without tuning them. Following the common practice in [2, 3, 6, 12, 13], we perform a cross-validation to choose other hyperparameters, *e.g.*, $\lambda_{\text{ALI}}$, $\lambda_{\text{KD}}$, and $\lambda_{\text{MEM}}$. We summarize in Table B hyperparameters for all experiments. On ADE20K [15], we perform a grid search to set the hyperparameters: $\lambda_{\text{ALI}} \in \{1, 2, 5\}$ and $\lambda_{\text{KD}} \in \{1, 10, 20\}$ for the first step; $\lambda_{\text{ALI}} \in \{0, 1, 2, 5\}$ and $\lambda_{\text{MEM}} \in \{0.1, 0.5, 1, 2\}$ for the third step. In particular, the grid search for the first step is performed only once at the first incremental stage (*i.e.*, $t = 2$), since the search on ADE20K is computationally expensive. In the subsequent stages, we use the same values of hyperparameters. We however perform a grid search at every incremental stage for the third step. This is because we fine-tune a classifier only for a single epoch, which is computationally acceptable. On PASCAL VOC [7], we have empirically found that a hIoU score on the cross-validation set decreases during training, when ISS models are trained with 30 epochs. A plausible reason is that the number of training samples on PASCAL VOC is relatively small. For example, the number of training samples on 20-1(1) of PASCAL VOC is 548, while the number of training samples on 100-50(1) of ADE20K is 9,390. We thus use a grid search to set the hyperparameters for the first step with $\lambda_{\text{ALI}} \in \{1, 2, 3, 5\}$, $\lambda_{\text{KD}} \in \{1, 10\}$, and the number of epochs (EP) $\in \{5, 10\}$. Since the search cost is relatively mild on PASCAL VOC, we perform the grid search at every incremental stage. For the third step, we also perform a grid search with $\lambda_{\text{ALI}} \in \{1, 2, 3, 5\}$ and $\lambda_{\text{MEM}} \in \{1, 2, 10, 20\}$.

## C More discussions

In this section, we first provide an analysis on ALI. Second, we extend the ablation studies in Tables 5 and 6 of the main paper in order to validate our design choices. Finally, we vary the number of memorized features to analyze its effect on performance, and empirically show that FAN [10] performs poorly when adopted in ISS.

**ALI.** Our ALI can be interpreted as keeping a balance between the maximum logit value and the weighted average of logit values for previous categories adaptively (See Sec.3.2 in the main paper). We conjecture that ALI helps a current model to obtain a more balanced classifier in terms of the norms of classifier weights. To empirically validate this, we show in Fig. C the difference between norms of classifier weights averaged over new and previous categories, denoted by $\|w\|_{\text{new}}$ and $\|w\|_{\text{prev}}$, respectively, during training. For visualization, we train each method on 16-5(1) of PASCAL VOC [7]. From this figure, we can see that applying the CE term alone fails to minimize the difference between $\|w\|_{\text{new}}$ and $\|w\|_{\text{prev}}$ during training. We can also see that both KD and CKD terms are not sufficient to minimize the difference, when being used with the CE term. On the other hand, using the

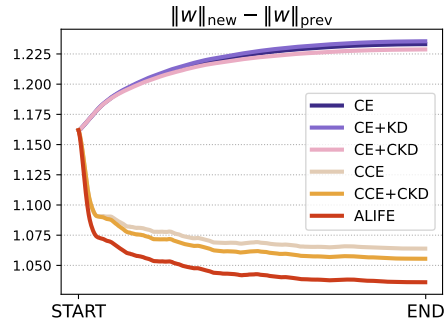

Figure C: $\|w\|_{\text{new}}$ and $\|w\|_{\text{prev}}$ indicate the norms of classifier weights averaged over new and previous categories, respectively. We plot the difference between $\|w\|_{\text{new}}$ and $\|w\|_{\text{prev}}$ during training.

CCE term alone reduces the difference during training, once again verifying that CCE alleviates the overfitting problem. Our approach also minimizes the difference consistently, achieving the lowest difference. This further strengthens the effectiveness of ALI and explains the reason why ALIFE outperforms MiB [2] in Tables 3 and 4 of the main paper.

Table C: Comparison of IoU scores using different loss terms of our approach in the first step. *Labeled*, *Unlabeled*, and *All* indicate that KD is applied for labeled, unlabeled, and all regions, respectively. Numbers in bold are the best performance, while underlined ones are the second best.

| Scenarios | CE | ALI | KD Labeled | KD Unlabeled | KD All | mIoU$_{base}$ | mIoU$_{new}$ | mIoU | hIoU |
|---|---|---|---|---|---|---|---|---|---|
| | ✓ | | | | | 16.45 | 6.94 | 14.19 | 9.74 |
| | ✓ | ✓ | | | | 75.50 | 49.81 | 69.39 | 60.02 |
| 16-5(1) | ✓ | ✓ | ✓ | | | **77.18** | **52.52** | **71.31** | **62.50** |
| | ✓ | ✓ | | ✓ | | 75.45 | 49.75 | 69.33 | 59.97 |
| | ✓ | ✓ | | | ✓ | <u>76.25</u> | <u>50.99</u> | <u>70.24</u> | <u>61.12</u> |
| | ✓ | | | | | 64.63 | 1.11 | 61.61 | 2.19 |
| | ✓ | ✓ | | | | 76.43 | 48.74 | 75.11 | 59.52 |
| 20-1(1) | ✓ | ✓ | ✓ | | | **76.61** | **49.36** | **75.31** | **60.03** |
| | ✓ | ✓ | | ✓ | | 76.48 | 48.90 | 75.17 | 59.65 |
| | ✓ | ✓ | | | ✓ | <u>76.51</u> | <u>49.07</u> | <u>75.20</u> | <u>59.79</u> |

Table D: Comparison of IoU scores using different loss terms of our approach in the third step. *Labeled*: CE or FL is applied only for labeled regions. *All\**: To apply CE or FL for all regions, we mark unlabeled regions as predictions of a previous model on-the-fly. Numbers in bold are the best performance, while underlined ones are the second best.

| Scenarios | CE Labeled | CE All* | FL Labeled | FL All* | ALI | MEM | mIoU$_{base}$ | mIoU$_{new}$ | mIoU | hIoU |
|---|---|---|---|---|---|---|---|---|---|---|
| | ✓ | | | | | | 76.11 | 48.03 | 69.42 | 58.89 |
| | | ✓ | | | | | 76.71 | 50.37 | 70.44 | 60.81 |
| | | | ✓ | | | | 76.51 | 49.70 | 70.13 | 60.26 |
| 16-5(1) | | | | ✓ | | | 76.81 | 50.98 | 70.66 | 61.29 |
| | | | | ✓ | | ✓ | <u>77.24</u> | <u>54.90</u> | <u>71.92</u> | <u>64.17</u> |
| | | | | ✓ | ✓ | | 77.25 | 52.88 | 71.44 | 62.78 |
| | | | | ✓ | ✓ | ✓ | **77.66** | **55.27** | **72.33** | **64.57** |
| | ✓ | | | | | | 76.40 | 38.33 | 74.59 | 50.97 |
| | | ✓ | | | | | 76.46 | 46.20 | 75.02 | 57.59 |
| | | | ✓ | | | | 76.38 | 43.27 | 74.80 | 55.22 |
| 20-1(1) | | | | ✓ | | | 76.42 | 46.63 | 75.01 | 57.91 |
| | | | | ✓ | | ✓ | <u>76.56</u> | 47.80 | 75.19 | 58.85 |
| | | | | ✓ | ✓ | | **76.72** | <u>52.23</u> | <u>75.55</u> | <u>62.15</u> |
| | | | | ✓ | ✓ | ✓ | **76.72** | **52.29** | **75.56** | **62.19** |

**More ablation studies.** Here we show more ablation studies including an analysis on our design choices. We report in Table C IoU scores using different loss terms of our approach in the first step. Note that this table contains the results of Table 5 in the main paper. From both scenarios, we can see that additionally using the KD term for labeled regions is beneficial to improving the performance. Table D compares performance using different loss terms of our approach in the third step, while including the results of Table 6 in the main paper. We can see in the first four rows of each scenario that the pseudo labeling strategy improves the performance. In particular, using the FL term along with the pseudo labeling strategy shows decent results. We can also see in the last three rows of each scenario that both ALI and MEM terms bring substantial IoU gains.

**Discussion on a feature replay scheme.** To analyze the effects of the number of memorized features, we vary the preset number $S$, and show in Table E performance w.r.t $S$. From this table, we can see that our approach to memorizing only 100 features for each previous category already shows decent results on both scenarios. The fourth row of each scenario shows that an approach to using features without updating them degrades the performance. This is natural in that the features extracted in the previous stage are not compatible with a classifier at a current stage. We empirically verify in the last row of each scenario that simply adopting FAN [10] in ISS shows sub-optimal performance. For a fair comparison, we train FAN with using Eq. (7) in the paper [10] with carefully tuning hyperparameters. After training, FAN updates saved features to fine-tune a classifier as in Eq. (19) of the main paper.

Table E: Comparison of IoU scores varying the preset number $S$ on PASCAL VOC [7]. No Update: we exploit features extracted from the previous stage to fine-tune a classifier at a current stage. That is, we do not update the features in this case.

| Scenarios | Methods | $S$ | mIoU$_{base}$ | mIoU$_{new}$ | mIoU | hIoU |
|---|---|---|---|---|---|---|
| | ALIFE-M | 100 | **77.68** | <u>55.22</u> | **72.33** | <u>64.54</u> |
| 16-5(1) | | 500 | **77.68** | <u>55.22</u> | **72.33** | <u>64.54</u> |
| | | 1000 | <u>77.66</u> | **55.27** | **72.33** | **64.57** |
| | No Update | 1000 | 72.90 | 55.16 | 68.67 | 62.79 |
| | FAN [10] | 1000 | 76.79 | 52.26 | 70.95 | 62.19 |
| | ALIFE-M | 100 | 76.70 | 52.28 | 75.54 | 62.18 |
| 20-1(1) | | 500 | **76.72** | **52.29** | **75.56** | **62.19** |
| | | 1000 | **76.72** | **52.29** | **75.56** | **62.19** |
| | No Update | 1000 | 76.64 | 52.09 | 75.47 | 62.03 |
| | FAN [10] | 1000 | 74.78 | 49.96 | 73.60 | 59.90 |

7

Table F: Comparison of IoU scores in the disjoint setting of PASCAL VOC [7]. We show standard deviations in parentheses. All numbers for other methods are copied from SSUL [3]. Numbers in bold are the best performance, while underlined ones are the second best.

| Scenarios | Methods | mIoU$_{base}$ | mIoU$_{new}$ | mIoU | hIoU |
|---|---|---|---|---|---|
| **16-5(1)** | MiB [2] | <u>71.80</u> | 43.30 | <u>64.70</u> | 54.02 |
| | PLOP [6] | 71.00 | 42.82 | 64.29 | 53.42 |
| | SSUL [3] | **76.44** | **45.60** | **69.10** | **57.12** |
| | ALIFE | 70.66 (0.64) | <u>44.03</u> (0.69) | 64.32 (0.42) | <u>54.25</u> (0.45) |
| **20-1(1)** | MiB [2] | 69.60 | 25.60 | 67.40 | 37.43 |
| | PLOP [6] | 75.37 | <u>38.89</u> | 73.64 | <u>51.31</u> |
| | SSUL [3] | **77.38** | 22.43 | <u>74.76</u> | 34.78 |
| | ALIFE | <u>76.31</u> (0.28) | **50.94** (2.14) | **75.11** (0.35) | **61.08** (1.61) |

**Discussion on the incremental gains from memorizing images or features on ADE20K.** We have observed that the gains from memorizing images or features in Table 3 of the main paper are relatively lower than those in Table 4. We speculate a plausible reason for this issue as follows. It is obvious that ADE20K [15] is more challenging than PASCAL VOC [7]. In particular, ADE20K contains 35 stuff categories, while PASCAL VOC has a single background category. Since 1) the background category always belongs to base categories, and 2) most unlabeled pixels on PASCAL VOC belong to the background category during incremental stages, the unlabeled pixels are less likely to contain future categories even in the overlapped setting. By contrast, on ADE20K, the single background category is split into multiple stuff categories that could possibly belong to future categories. SSUL-M [3] memorizes previously seen images, which contain unlabeled regions, together with ground-truth labels. Since pixels of stuff categories occupy about 60% of all the pixels on ADE20K (See Sec. 4 in [15]), the unlabeled regions could increase when the stuff categories belong to future ones. This might be problematic in that the number of labeled pixels decreases accordingly, lessening the effectiveness of the memoized images. In our case, the feature alignment scheme computes the correlation score between $f^{t-1}(\mathbf{p})$ and $m_c(s)$ (See Eq. (12) in the main paper). If the label of position belongs to future categories, it might result in erroneous correlations, reducing the quality of feature alignment.

## D  More results

**Disjoint setting.** In the disjoint setting, we assume that (1) all categories even including future ones are known in advance and (2) training samples of a current dataset do not contain any categories that would be seen in the future. We report in Table F results of our approach in the disjoint setting of PASCAL VOC [7]. From this table, we can see that ALIFE shows better results than MiB [2] and PLOP [6] in terms of hIoU scores on both 16-5(1) and 20-1(1) cases. In particular, ALIFE outperforms SSUL [3], which exploits an off-the-shelf saliency detector [9], by a large margin in terms of hIoU scores on 20-1(1). This once again demonstrates the effectiveness of our ALI. Note that the disjoint setting is quite different from an overlapped setting in the main paper, where training samples of the current dataset can contain any categories. As pointed out in previous works [2, 3, 6], the overlapped setting is more practical and challenging.

**Qualitative results.** We show in Figs. D and E qualitative results on PASCAL VOC [7]. We show in the last row of each figure failure cases. From the third and fourth columns of Fig. D, we can see that MiB [2] and PLOP [6] struggle to preserve the discriminability for previous categories, *e.g.*, cow, table, and chair. In particular, both methods misclassify cow and chair categories as sheep and sofa, respectively, due to the overfitting problem. ALIFE and ALIFE-M alleviate this problem, showing better segmentation results. We can also see from the last two columns of Fig. D that replaying features gives better results than ALIFE. For example, the fourth row shows that ALIFE-M produces accurate predcitions on the regions for the cow category, where results of ALIFE are incorrect. Figure E also shows a qualitative comparison of ours with MiB and PLOP. Again, the results of MiB and PLOP show that both methods are prone to overfitting to new categories (*e.g.*, a tv category in

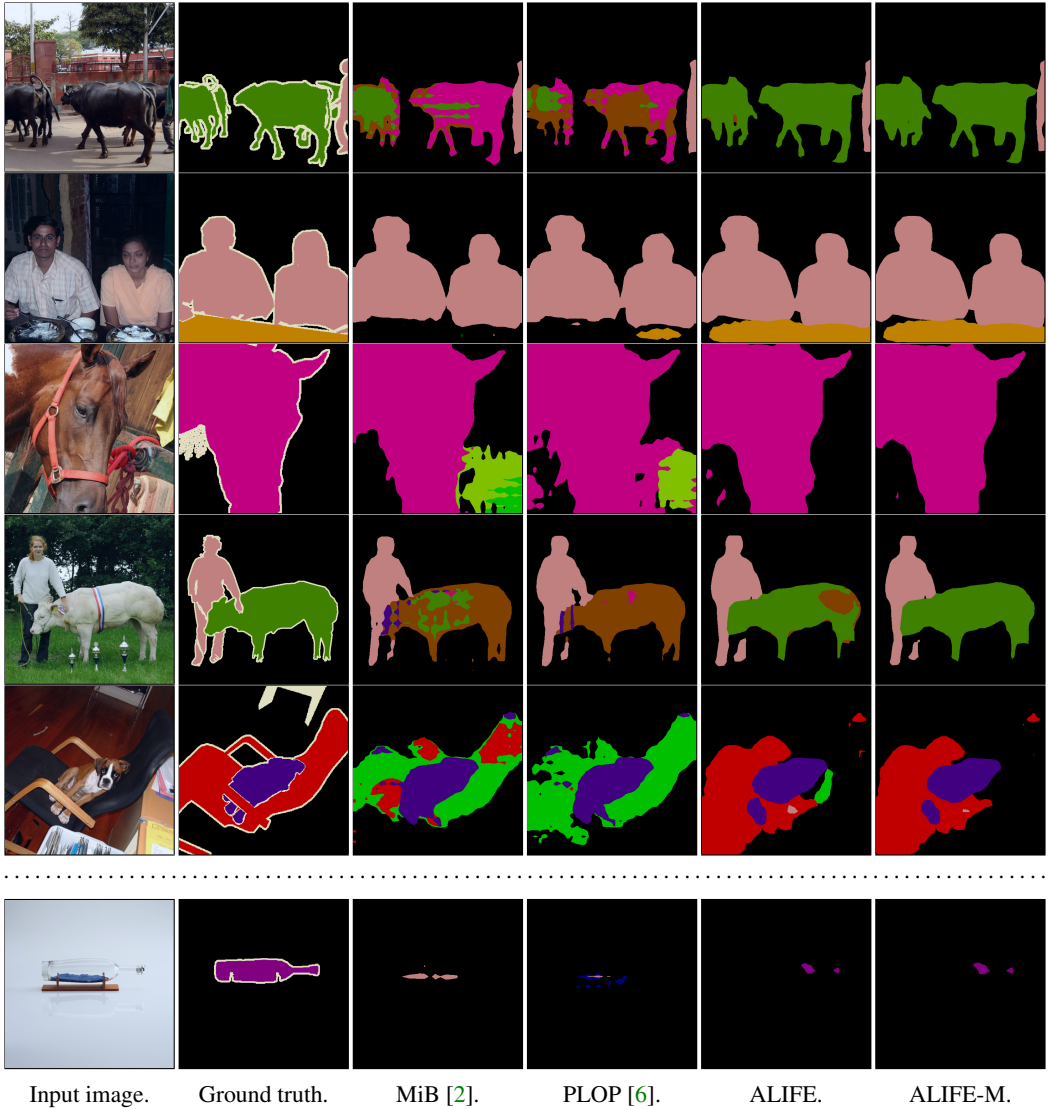

| Input image. | Ground truth. | MiB [2]. | PLOP [6]. | ALIFE. | ALIFE-M. |

Figure D: Visual comparison of ours and other methods [2, 6] on 16-5(1) for the PASCAL VOC [7] validation set. Each method is trained to recognize 5 novel categories (*i.e.* potted plant, sheep, sofa, train, and tv) after learning 16 categories. The last row shows a failure case. Best viewed in color.

this case). We can see that ALIFE already shows decent results and ALIFE-M further improves the quality of segmentation results.

# References

[1] Samuel Rota Bulo, Lorenzo Porzi, and Peter Kontschieder. In-place activated batchnorm for memory-optimized training of dnns. In *CVPR*, 2018.

[2] Fabio Cermelli, Massimiliano Mancini, Samuel Rota Bulo, Elisa Ricci, and Barbara Caputo. Modeling the background for incremental learning in semantic segmentation. In *CVPR*, 2020.

[3] Sungmin Cha, YoungJoon Yoo, Taesup Moon, et al. SSUL: Semantic segmentation with unknown label for exemplar-based class-incremental learning. *NeurIPS*, 2021.

[4] Liang-Chieh Chen, George Papandreou, Florian Schroff, and Hartwig Adam. Rethinking atrous convolution for semantic image segmentation. *arXiv*, 2017.

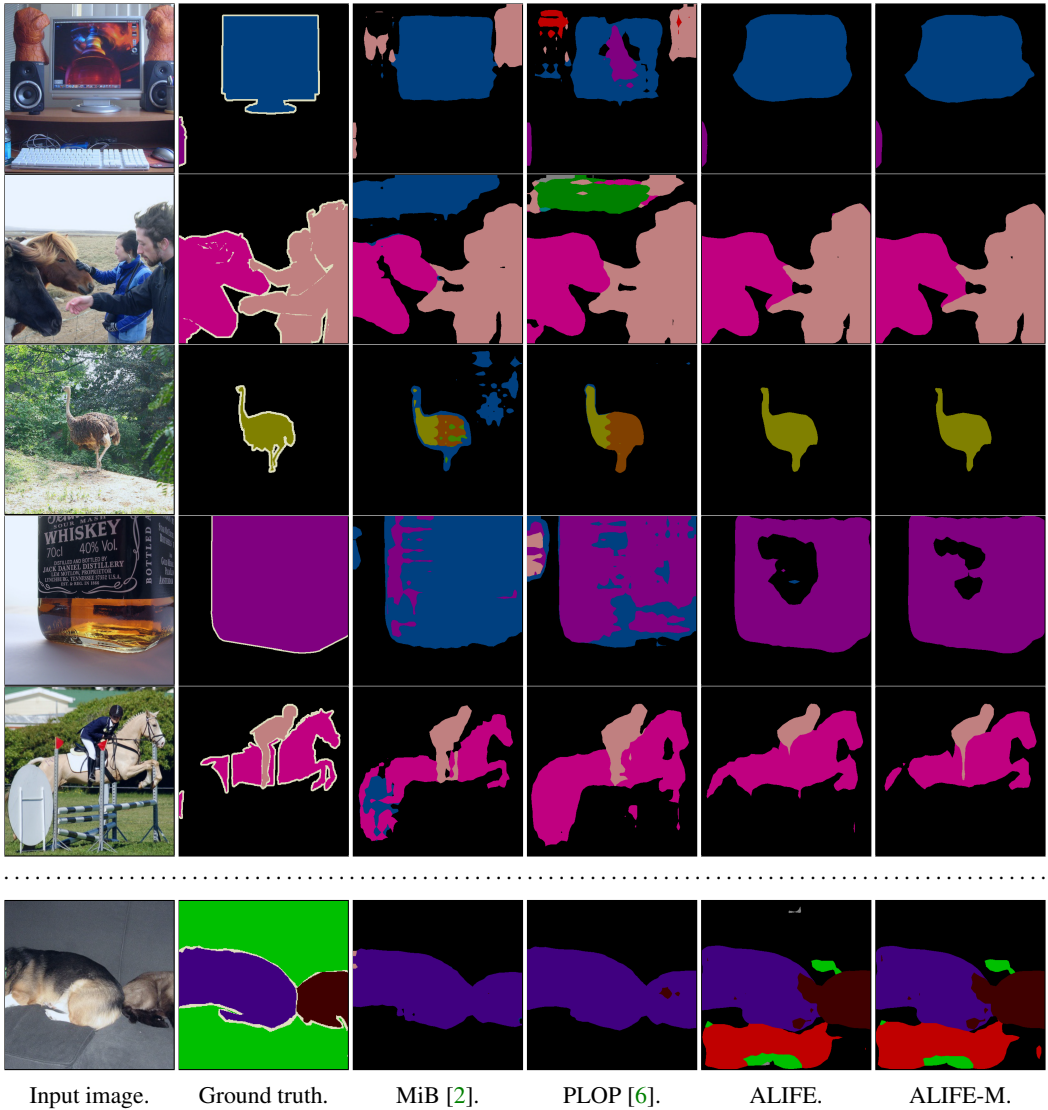

Figure E: Visual comparison of ours and other methods [2, 6] on 20-1(1) for the PASCAL VOC [7] validation set. Each method is trained to recognize a tv category after learning 20 categories. The last row shows a failure case. Best viewed in color.

[5] Jia Deng, Wei Dong, Richard Socher, Li-Jia Li, Kai Li, and Li Fei-Fei. ImageNet: A large-scale hierarchical image database. In *CVPR*, 2009.

[6] Arthur Douillard, Yifu Chen, Arnaud Dapogny, and Matthieu Cord. PLOP: Learning without forgetting for continual semantic segmentation. In *CVPR*, 2021.

[7] Mark Everingham, Luc Van Gool, Christopher KI Williams, John Winn, and Andrew Zisserman. The PASCAL visual object classes (VOC) challenge. *IJCV*, 88(2):303–338, 2010.

[8] Kaiming He, Xiangyu Zhang, Shaoqing Ren, and Jian Sun. Deep residual learning for image recognition. In *CVPR*, 2016.

[9] Qibin Hou, Ming-Ming Cheng, Xiaowei Hu, Ali Borji, Zhuowen Tu, and Philip HS Torr. Deeply supervised salient object detection with short connections. In *CVPR*, 2017.

[10] Ahmet Iscen, Jeffrey Zhang, Svetlana Lazebnik, and Cordelia Schmid. Memory-efficient incremental learning through feature adaptation. In *ECCV*, 2020.

[11] Tsung-Yi Lin, Priya Goyal, Ross Girshick, Kaiming He, and Piotr Dollár. Focal loss for dense object detection. In *ICCV*, 2017.

[12] Andrea Maracani, Umberto Michieli, Marco Toldo, and Pietro Zanuttigh. RECALL: Replay-based continual learning in semantic segmentation. In *ICCV*, 2021.

[13] Umberto Michieli and Pietro Zanuttigh. Continual semantic segmentation via repulsion-attraction of sparse and disentangled latent representations. In *CVPR*, 2021.

[14] Adam Paszke, Sam Gross, Soumith Chintala, Gregory Chanan, Edward Yang, Zachary DeVito, Zeming Lin, Alban Desmaison, Luca Antiga, and Adam Lerer. Automatic differentiation in pytorch. In *NeurIPS Workshop*, 2017.

[15] Bolei Zhou, Hang Zhao, Xavier Puig, Sanja Fidler, Adela Barriuso, and Antonio Torralba. Scene parsing through ADE20K dataset. In *CVPR*, 2017.

**Algorithm 1** Pseudo code of incremental semantic segmentation with ALIFE.

```
 1: // We omit a training process at a base stage (t = 1)
 2: for t ∈ {2, ..., T} do
 3:     // Step 1
 4:     {ϕ^t, w^t} ← {ϕ^{t-1}, w^{t-1}}  // Initialize a current model
 5:     ep ← 0
 6:     repeat
 7:         Sample a mini-batch B ∼ D^t  // B indicates a mini-batch
 8:         Update network weights of the current model {ϕ^t, w^t} ← SGD(B, {ϕ^t, w^t}, L_{S1})  // L_{S1} in Eq. (10)
 9:         ep ← ep + 1
10:     until ep=EP  // EP indicates the number of training epochs
11:
12:     // Step 2
13:     // Extract features which are used to replay in subsequent stages
14:     Freeze {ϕ^t, w^t}
15:     for c ∈ C^t_{new} do
16:         T^t_c ← [ ]  // T^t_c indicates a set of features for the category c
17:         s ← 0
18:         repeat
19:             (x, y) ∼ D^t
20:             Extract a feature map  f^t ← ϕ^t(x)
21:             Average features for the category c   m_c ← (1/|R_c|) ∑_{p∈R_c} f^t(p)
22:             T^t_c ← [T^t_c; m_c]
23:             s ← s + 1
24:         until s = S  // S indicates the preset number
25:     end for
26:
27:     // Train rotation matrices
28:     Freeze {ϕ^t, w^t}
29:     ep ← 0
30:     R ← {R_c | c ∈ C^t_{prev}}  // Initialize a set of rotation matrices randomly (See Eqs. (xi) and (xii))
31:     repeat
32:         Sample a mini-batch B ∼ D^t
33:         Update parameters of the rotation matrices R ← SGD(B, R, L_{S2})  // L_{S2} in Eq. (15)
34:         ep ← ep + 1
35:     until ep=10
36:
37:     // Step 3
38:     // Update saved features which are extracted in the previous stage
39:     Freeze R
40:     for c ∈ C^t_{prev} do
41:         // M^{t-1}_c indicates a set of memorized features for the category c in the previous stage
42:         Rotate features M̂^{t-1}_c ← R_c M^{t-1}_c
43:     end for
44:
45:     // Fine-tune a classifier
46:     Freeze ϕ^t
47:     M̂^{t-1} ← {M̂^{t-1}_c | c ∈ C^t_{prev}}
48:     ep ← 0
49:     repeat
50:         Sample a mini-batch B ∼ D^t
51:         Update classifier weights of the current model   w^t ← SGD({B, M̂^{t-1}}, w^t, L_{S3})  // L_{S3} in Eq. (19)
52:         ep ← ep + 1
53:     until ep=1
54:
55:     // Concatenate rotated features M̂^{t-1}_c and extracted features T^t_c
56:     for c ∈ C^t_{all} do
57:         if c ∈ C^t_{prev} then
58:             M^t_c ← M̂^{t-1}_c
59:         end if
60:         if c ∈ C^t_{new} then
61:             M^t_c ← T^t_c
62:         end if
63:     end for
64: end for
```



\end{document}